\documentclass{article}

 \usepackage[preprint]{neurips_2026}

\usepackage[utf8]{inputenc} 
\usepackage[T1]{fontenc}    
\usepackage{hyperref}       
\usepackage{url}            
\usepackage{booktabs}       
\usepackage{amsfonts}       
\usepackage{nicefrac}       
\usepackage{microtype}      
\usepackage{xcolor}         

\usepackage{graphicx}
\usepackage{enumitem}
\usepackage{pifont}         
\usepackage{defs}
\usepackage{amsmath}
\usepackage[capitalise]{cleveref}
\usepackage{algorithm}
\usepackage{algpseudocode}

\title{SubsurfaceGen: Procedural Generation of Field-Scale Earth Models and Seismic Data}

\author{%
  Joseph Stitt, Pratik Rathore, Madeleine Udell, Ching-Yao Lai \\
  Stanford University \\
  \texttt{jdstitt@sep.stanford.edu,\{pratikr,udell,cyaolai\}@stanford.edu}
}

\begin{document}

\maketitle
\vspace{-2.5em}
{\centering
    \small
    \github{}~\textbf{SubsurfaceGen}: \href{https://anonymous.4open.science/r/subsurfacegen-8961}{https://anonymous.4open.science/r/subsurfacegen-8961} \\
    \github{}~\textbf{Experiments}: \href{https://anonymous.4open.science/r/subsurfacegen-experiments-4C98}{https://anonymous.4open.science/r/subsurfacegen-experiments-4C98} \\
    \hflogo{}~\textbf{Field-Scale Dataset}: \href{https://huggingface.co/datasets/subsurfacegen/field-scale-dataset}{https://huggingface.co/datasets/subsurfacegen/field-scale-dataset} \\
    \hflogo{}~\textbf{Field-Scale Dataset (Preview)}: \href{https://huggingface.co/datasets/subsurfacegen/field-scale-dataset-preview}{https://huggingface.co/datasets/subsurfacegen/field-scale-dataset-preview}
    \par}

\begin{abstract}
    Full waveform inversion (FWI) is the gold standard for subsurface imaging, with applications from carbon sequestration to energy and mineral exploration to earthquake hazard assessment.
    Machine learning approaches to FWI need field-scale, geologically diverse, and physically realistic training data, but existing resources such as Marmousi, SEAM, and OpenFWI fall short on spatial extent, temporal extent, geological diversity, and physical realism.
    We address these limitations with SubsurfaceGen, a GPU-accelerated generator for 3D velocity models and seismic data.
    Along with SubsurfaceGen, we release a paired dataset of 4{,}276 2D velocity slices, 5\,s wavefields, and 8\,s shot gathers drawn from 42 realistic, field-scale 3D velocity models, each spanning $10\,\text{km} \times 10\,\text{km}$ laterally and $6.19\,\text{km}$ deep at $10\,\text{m}$ resolution.
    The dataset spans six geological settings---four built with SubsurfaceGen and two drawn from prior sources---relevant for carbon sequestration and hydrocarbon exploration.
    We use this dataset to evaluate neural operators on wavefield prediction and encoder--decoders on end-to-end velocity inversion, holding out one geological setting for out-of-distribution testing.
    These experiments surface failure modes at field-scale and demonstrate how SubsurfaceGen and the associated dataset can impact ML-based FWI.
\end{abstract}


\section{Introduction}
\label{sec:introduction}


Subsurface imaging---the task of reconstructing the Earth's interior from indirect surface measurements---carries significant economic and societal consequences, from verifying the integrity of carbon sequestration sites critical for addressing climate change, to discovering energy and mineral resources that underpin modern infrastructure, to assessing earthquake hazards that shape emergency planning in densely populated regions.
The gold standard for subsurface imaging is full waveform inversion (FWI) \citep{lailly1983seismic,tarantola1984inversion,virieux2009overview}, which reconstructs a \textit{velocity model}, a spatial map of acoustic wave speed in the subsurface, from data collected in a \textit{seismic survey}, in which an acoustic source is fired at many positions across a region of interest and an array of receivers records the reflections from the subsurface.
Each source firing produces a \textit{shot gather}: the time series recorded at the receivers in response to that firing.
FWI matches these recordings to \textit{wavefields} simulated by solving the wave equation through a candidate velocity model.
However, FWI is notoriously difficult: it requires the solution of a high-dimensional, non-convex, PDE-constrained optimization problem, which suffers from cycle skipping (i.e., bad local minima) \citep{virieux2009overview,yao2019tackling}, sensitivity to the initial model \citep{virieux2009overview}, and computational costs that can scale to millions of core hours for realistic 3D surveys \citep{schiemenz2013accelerated}.
Decades of research have produced sophisticated regularization schemes \citep{symes2008migration,biondi2014simultaneous,barnier2023practical} and multiscale strategies \citep{bunks1995multiscale,fichtner2011full}, but the difficulties of FWI remain far from solved.

Machine learning is promising on several of these fronts.
Neural operators can accelerate PDE solves in FWI, reducing overall computational costs, while maintaining accuracy and reliability \citep{yang2021seismic, yang2023unoseismic, zhang2023elasticfno, huang2025scattered}.
CNN-based encoder--decoders can directly map seismic measurements to velocity models, sidestepping the cycle skipping, initialization sensitivity, and computational costs of traditional FWI methods \citep{arayapolo2018dltomography, yang2019fcnvmb, wu2020inversionnet, zhang2020velocitygan, farris2023learning, wang2023svit}.
Generative models can learn data-driven regularizers from existing velocity models, capturing realistic geological features that classical regularizers like Tikhonov and total variation are unable to represent \citep{mosser2020ganprior, stitt2023deepdix, wang2023priorfwi, stitt2025latentdiff}.

However, progress on these fronts is bottlenecked by data.
The velocity models used to train and evaluate ML-based FWI should resemble the surveys they will be deployed on---\textit{field-scale}, geologically diverse, and physically realistic---and should be extendable.
By field-scale, we mean velocity models that match the geometries of real surveys: tens of kilometers laterally, several kilometers deep, with recordings over several seconds.
By extendable, we mean the ability to generate new velocity models on demand.
Field-scale data is necessary because cycle skipping, illumination gaps, and low-frequency recovery worsen with spatial extent, and because imaging depth scales with recording time: carbon storage and hydrocarbon reservoirs sit several kilometers below the surface, requiring recordings of 5 s or more to image.
Geological diversity is needed to study generalization to unseen geologies.
Physical realism guards against ML methods producing nonsensical geological features.
Extendability supports studying scaling behavior, distribution shift, and generalization across new geological settings.

Popular datasets for FWI, such as Marmousi \citep{versteeg1994marmousi,martin2006marmousi2}, SEAM \citep{fehler2011seam}, and OpenFWI \citep{deng2022openfwi}, fall short on at least one of these criteria, whereas SubsurfaceGen, the data generator we introduce in this paper, satisfies all four (\cref{tab:resource_comparison}).

\begin{table}[h]
    \small
    \centering
    \caption{SubsurfaceGen vs. existing datasets for FWI; see also \cref{subsec:popular_datasets}.}
    \label{tab:resource_comparison}
    \begin{tabular}{l c c c c}
        \toprule
        \textbf{Resource}    & \textbf{Field-scale?} & \textbf{Geologically diverse?} & \textbf{Physically realistic?} & \textbf{Extendable?} \\
        \midrule
        Marmousi             & \cmark                & \xmark                         & \cmark                         & \xmark               \\
        SEAM                 & \cmark                & \xmark                         & \cmark                         & \xmark               \\
        OpenFWI              & \pmark                & \pmark                         & \xmark                         & \xmark               \\
        SubsurfaceGen (ours) & \cmark                & \cmark                         & \cmark                         & \cmark               \\
        \bottomrule
    \end{tabular}
\end{table}

\begin{figure}[t]
    \centering
    \includegraphics[width=\linewidth]{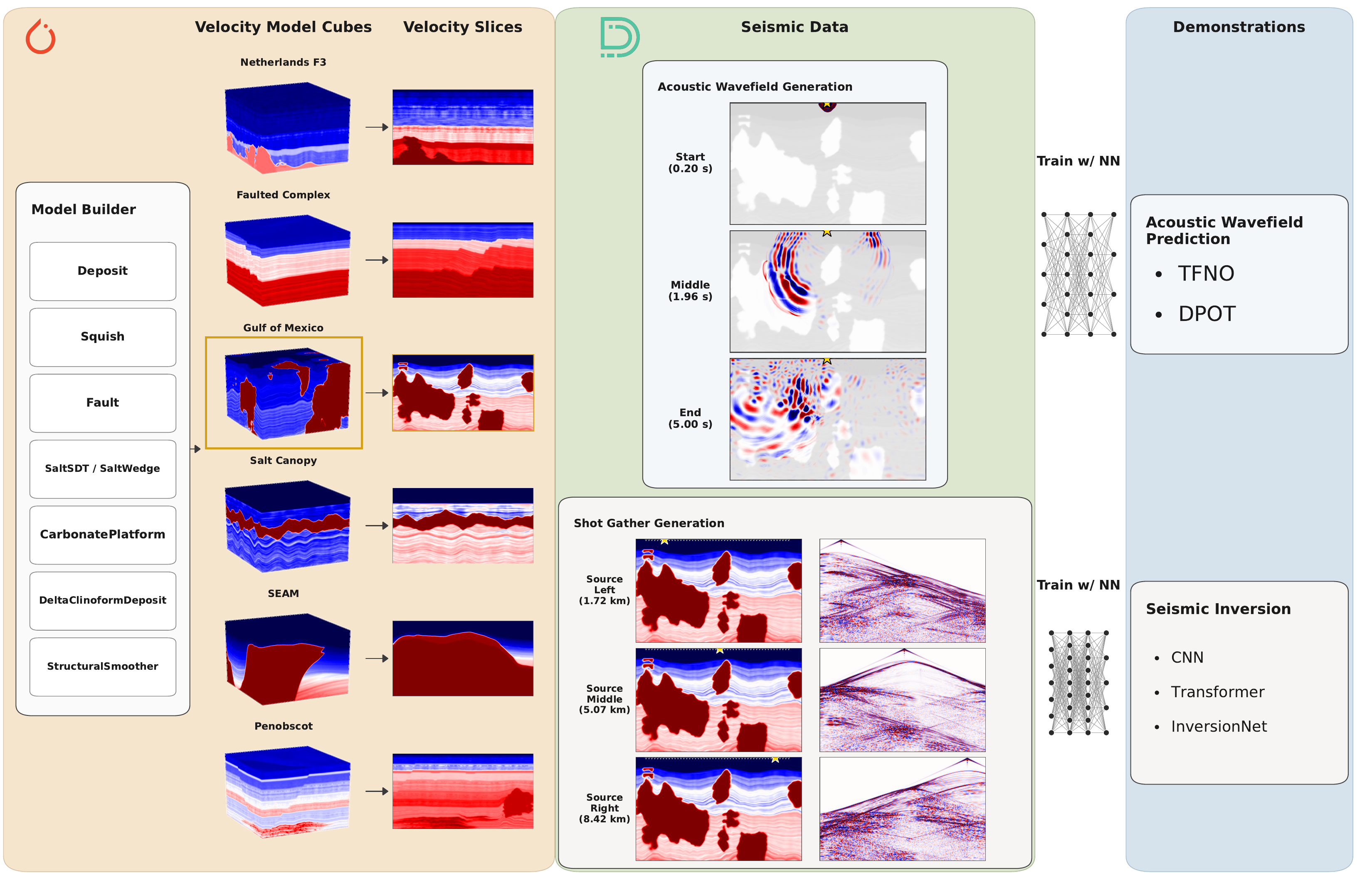}
    \caption{SubsurfaceGen is a GPU-accelerated velocity model builder (using PyTorch) and seismic data generator (using \devito), which can be used to produce training data for ML-based FWI.
        We use SubsurfaceGen to create a dataset of 42 realistic, field-scale 3D velocity models along with (velocity slice, wavefield) and (velocity model, shot gather) pairs.
        We use these pairs to train neural operators for wavefield prediction and encoder--decoders for end-to-end inversion from shot gathers.
    }
    \label{fig:overview}
\end{figure}

SubsurfaceGen is a state-of-the-art data generator that enables procedural generation of field-scale, geologically diverse, physically realistic 3D velocity models and seismic data.
Our contributions (illustrated in \cref{fig:overview}) are as follows:

\begin{itemize}
    \item \textbf{Realistic velocity model generation (\cref{sec:model_builder}).} SubsurfaceGen allows users to deposit geological layers with interbeds (alternating layers of contrasting velocity), and add folding, faults, salt bodies, clinoforms, and carbonate platforms, together with structure-oriented smoothing \citep{hale2009structureoriented} to suppress numerical artifacts.
          To our knowledge, no other open-source software supports procedural generation of this range of features.
    \item \textbf{End-to-end seismic data generation (\cref{sec:seismic_generation}).} SubsurfaceGen generates shot gathers and wavefields from 2D slices of 3D velocity models using \devito \citep{louboutin2019devito}, providing paired (velocity model, seismic data) samples for training and evaluation.
          Both velocity model and seismic data generation are GPU-accelerated, with model generation achieving up to 26.8$\times$ speedup over CPU.
    \item \textbf{Example 4,276 sample field-scale dataset (\cref{subsec:model_generation_huggingface,subsec:seismic_generation_huggingface}).} We release a dataset on \huggingface containing 42 field-scale 3D velocity models (each 10 km $\times$ 10 km $\times$ 6.19 km).
          From these we extract 4,276 2D slices, each paired with seismic data (5 s wavefields, 8 s shot gathers).
          The dataset spans six geological settings: four generated with SubsurfaceGen (based on basins in the North Sea, Gulf of Mexico, and Nova Scotia, plus a model with a large number of faults) and two drawn from prior sources (a model from a legacy 3D velocity model builder and the canonical SEAM model) to broaden geological coverage.
    \item \textbf{Field-scale experiments for ML-based FWI (\cref{sec:wavefield_prediction_experiments,sec:inversion_experiments}).} We demonstrate how SubsurfaceGen and the field-scale dataset impact ML-based FWI: for wavefield prediction, the field-scale grid forces predictions to be chunked, which motivates an adaptation of optimal checkpointing \citep{symes2007checkpointing}; for end-to-end inversion, geological diversity supports cross-geology generalization studies, opening a new possibility for evaluating architectures.
\end{itemize}

The rest of the paper proceeds as follows: \cref{sec:wave_equation} introduces the acoustic wave equation, \cref{sec:model_builder,sec:seismic_generation} describe SubsurfaceGen's velocity model and seismic data generation components, \cref{sec:wavefield_prediction_experiments,sec:inversion_experiments} present our experiments for ML-based FWI, and \cref{app:related_work} covers related work.

\section{The Acoustic Wave Equation}
\label{sec:wave_equation}

The acoustic wave equation maps a velocity model $v_p(\mathbf{x})$ to a wavefield $p(\mathbf{x}, t)$ that would be recorded in a real-world seismic survey.
This mapping is central to FWI and the ML methods in \cref{sec:introduction}: FWI inverts it, neural operators approximate it, and encoder--decoders for end-to-end inversion are trained against data produced by it.
SubsurfaceGen solves the 2D acoustic, constant-density, isotropic form of the wave equation with a source term:
\begin{equation}
    \frac{1}{v_p(\mathbf{x})^2}\,\frac{\partial^2 p(\mathbf{x},t)}{\partial t^2}
    \;-\; \nabla^2 p(\mathbf{x},t)
    \;=\; s(\mathbf{x},t).
    \label{eq:acoustic}
\end{equation}
Here $\mathbf{x} = (x, z)$ is the spatial coordinate (lateral $x$, depth $z$), $t$ is time, $p(\mathbf{x}, t)$ is the wavefield, $v_p(\mathbf{x})$ is the velocity model, $\nabla^2 = \partial_x^2 + \partial_z^2$ is the Laplacian, and $s(\mathbf{x}, t)$ is the source term.
SubsurfaceGen models the source $s(\mathbf{x}, t)$ as a Ricker wavelet \citep{ricker1953form}, which is standard in seismic processing \citep{wang2015ricker}.
Initial and boundary conditions are deferred to \cref{app:seismic_forward}.

Solving \eqref{eq:acoustic} produces the wavefield $p(\mathbf{x}, t)$
across space and time.
However, the wavefield is not observed at all spatial locations in a seismic survey.
Seismic surveys place a sparse line of sensors at
the surface, called \textit{receivers}, that record the time history of
the wavefield.
SubsurfaceGen reproduces this with a sampling
operator $\Rc$ that observes the wavefield $p$ at the receiver
locations: the result is a shot gather $\Rc p$, a 2D tensor with axes for time
and receiver index.
The remaining challenge is generating diverse, realistic velocity models, which we address in \cref{sec:model_builder}.

\section{SubsurfaceGen: Procedural Velocity Model Generation}
\label{sec:model_builder}

SubsurfaceGen provides functionality for generating realistic, field-scale 3D velocity models using a catalog of modules that emulate geological features observed in real-world seismic data.
These modules include interbeds, faults, salt bodies, clinoforms, and carbonate platforms, which are essential for characterizing sedimentary settings that arise in carbon sequestration and hydrocarbon exploration.
We have written SubsurfaceGen's model building functionality in PyTorch, which enables GPU-accelerated velocity model generation, making it easier for practitioners to generate large datasets of velocity models.
We also describe our process for generating velocity models that are contained in the field-scale dataset on \huggingface.
Additional details are available in \cref{app:model_builder_modules,app:model_details,app:builder_benchmark}.

\subsection{Velocity Model Builder}

\paragraph{The module catalog.}
SubsurfaceGen introduces eight modules for building velocity models.
Three of them (\deposit, \squish, \fault) improve on the synthetic-model package from \citet{clapp2018syntheticjupyter, clapp2022mlsynthetic, sep2024syntheticmodel}.
The remaining five modules (\saltsdt, \saltwedge, \carbonateplatform, \deltaclinoformdeposit, \structuralsmoother) are original to SubsurfaceGen.
At a high level: \deposit adds stratified layers, \squish warps layers into folds, \fault cuts and shifts layers, \saltsdt emplaces salt bodies and \saltwedge deforms layers around them, \carbonateplatform builds reef structures, \deltaclinoformdeposit places deltaic formations, and \structuralsmoother applies geology-aware smoothing.
\cref{tab:module_summary} provides a full description of each module.

\begin{figure}[t]
    \centering
    \includegraphics[width=0.95\linewidth]{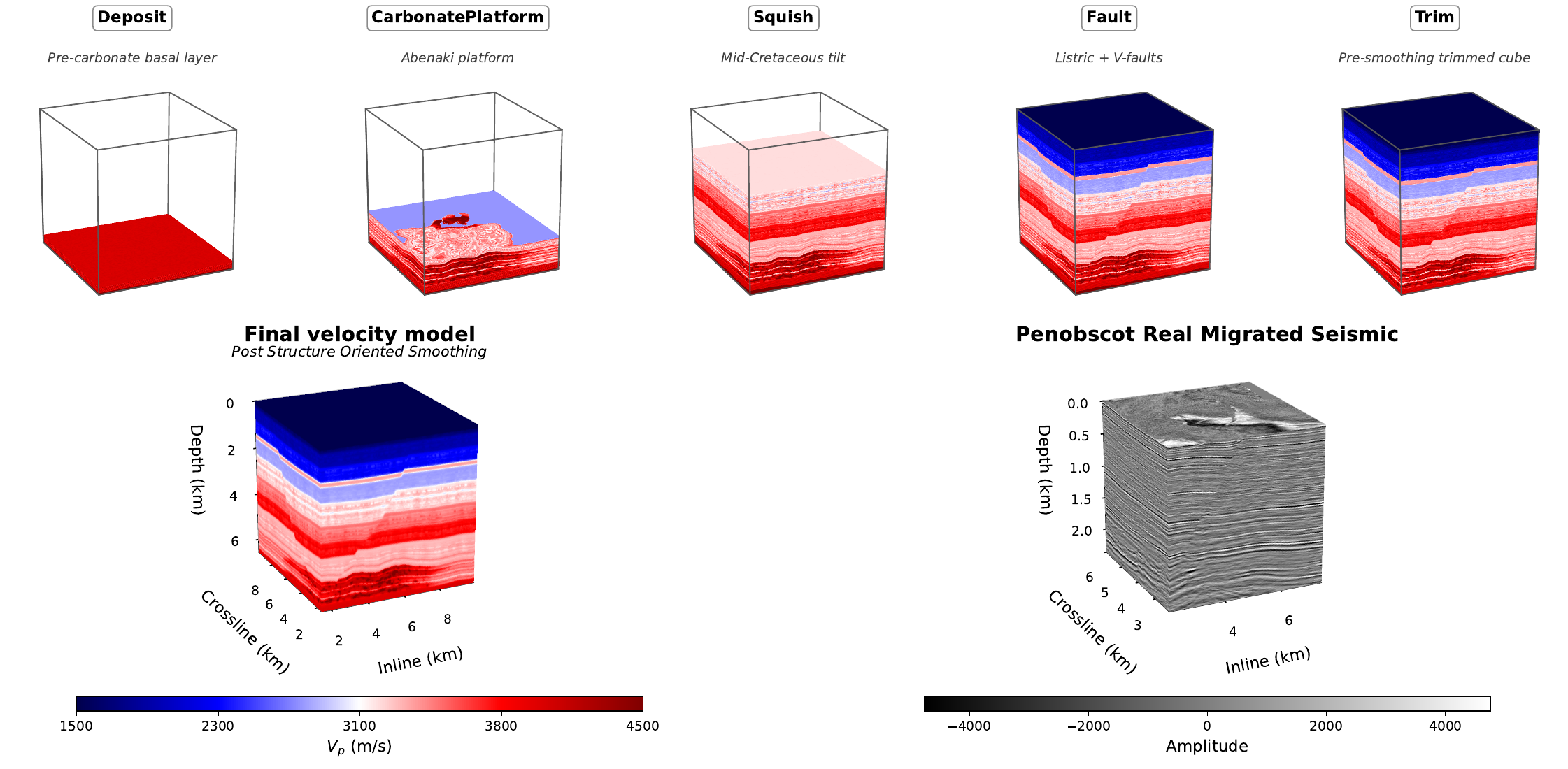}
    \caption{Top: Step-by-step construction of a Penobscot velocity
        model using SubsurfaceGen modules (we omit some steps for simplicity).
        Bottom: The velocity model juxtaposed against the Penobscot migrated cube.
        The velocity model captures key geological features of the migrated cube.}
    \label{fig:penobscot_flow}
\end{figure}

\paragraph{Building velocity models.}
Velocity models are built by \textit{applying} the modules in sequence, starting from a flat, homogeneous volume (called the \textit{basement}) and adding modules going up to the surface.
Each module takes the current velocity model as input and modifies it according to its specific geological feature.
The order of module application follows geological time, with older features laid down before younger ones.
After all modules have been applied, the user can apply structure-oriented smoothing (SOS) \citep{hale2009structureoriented} via \structuralsmoother to remove numerical artifacts from the build process while preserving important geological features like faults and bed contacts.
The model building process is highly customizable, allowing users to specify parameters for each module to generate a variety of velocity models that capture the diversity of geological settings observed in real subsurface scenarios.
\cref{fig:penobscot_flow} illustrates this process for a Penobscot velocity model.

\paragraph{GPU generation makes SubsurfaceGen extensible.}
\cref{tab:builder_speedup} reports per-setting build times on GPU vs. CPU: GPU yields a 7.2$\times$--26.8$\times$ speedup, bringing build times down from 1--2 hours on CPU to 5--10 minutes on GPU.
This makes SubsurfaceGen particularly useful for practitioners who want to adapt it to a new geological setting or study scaling behavior.

\begin{table}[h]
    \centering
    \small
    \caption{Total build time per geological setting.
        GPU: one 80 GB NVIDIA A100.
        CPU: 16 cores of an AMD EPYC 7763 with 256 GB of RAM.
        A full breakdown of the timings is in \cref{app:builder_benchmark}.}
    \label{tab:builder_speedup}
    \begin{tabular}{lrrr}
        \toprule
        Setting        & GPU build (min) & CPU build (min) & Speedup      \\
        \midrule
        Penobscot      & 5.57            & 92.67           & 16.6$\times$ \\
        F3             & 11.49           & 82.29           & 7.2$\times$  \\
        Fault          & 5.84            & 156.57          & 26.8$\times$ \\
        Gulf of Mexico & 6.70            & 100.21          & 15.0$\times$ \\
        \bottomrule
    \end{tabular}
\end{table}

\subsection{Generation of Field-Scale Velocity Models for \huggingface Dataset}
\label{subsec:model_generation_huggingface}

\paragraph{Building the dataset with SubsurfaceGen.}
We use SubsurfaceGen to generate the velocity models in our \huggingface dataset, building each model by applying modules from basement to surface.
The dataset contains 42 field-scale velocity models in total, each $6.19\,\text{km} \times 10\,\text{km} \times 10\,\text{km}$ at 10\,m resolution, organized into six geological settings.
Four of the settings (Penobscot, F3, Gulf of Mexico, Fault) are SubsurfaceGen builds.
The remaining two are not: SEAM is an existing model \citep{fehler2011seam} and Salt Canopy was generated using a legacy model building code with inspiration from \citet{farris2023learning}.
A full inventory is given in \cref{tab:model_inventory} of \cref{app:model_details}.

\paragraph{Diversity between and within geological settings.}
Our velocity models are designed to mimic publicly-available \textit{migrated cubes}---estimates of subsurface structure reconstructed from seismic reflection data recorded at the surface \citep{lo1994seismictomography}, similar to the way in which medical CT scans estimate internal anatomy from X-ray attenuation measurements taken outside the body.
The velocity models in our dataset capture diversity across two levels: \textit{between geological settings} and \textit{within each setting}.
We provide six settings (Penobscot, F3, Gulf of Mexico, Fault, Salt Canopy, SEAM) so that downstream models see a wide variety of geological features.
Within each setting, we provide multiple realizations drawn from the same regional distribution (we describe our methodology in the next paragraph).
Publicly-available migrated cubes are scarce, so within-setting diversity allows us to leverage the few cubes that do exist while still providing enough velocity models for training.

\paragraph{Matching to migrated cubes and sampling variants.}
We match a velocity model for each setting to the corresponding migrated cube, then sample variants around the matched parameters.
The build process is the deterministic mapping $\Mc(\theta) = V$, where $\Mc$ denotes the velocity model builder, $\theta$ denotes the parameter vector configuring SubsurfaceGen modules, and $V$ denotes the resulting velocity model.
Matching is manual: given a public migrated cube $C_{\text{GT}}$, we pick a parameter vector $\theta$, generate $\Mc(\theta)$, compare to $C_{\text{GT}}$ by visual inspection, and iterate until $\Mc(\theta_{\text{cal}})$ resembles $C_{\text{GT}}$.
We then define a sampling distribution $\Dc$ centered at $\theta_{\text{cal}}$, draw
\begin{equation*}
    \theta_1, \ldots, \theta_k \stackrel{\text{iid}}{\sim} \Dc,
\end{equation*}
and produce the setting's models as $\Mc(\theta_{\text{cal}}), \Mc(\theta_1), \ldots, \Mc(\theta_k)$.
\cref{fig:penobscot_flow} shows the public migrated cube $C_{\text{GT}}$ alongside our matched synthetic $\Mc(\theta_{\text{cal}})$ for the Penobscot setting; per-setting geological details and build scripts are in \cref{app:model_details}.

\section{SubsurfaceGen: Seismic Data Generation}
\label{sec:seismic_generation}

We describe the seismic data generation functionality of SubsurfaceGen and explain how we use it to build the seismic data in our \huggingface dataset.
For each 3D velocity model from \cref{sec:model_builder}, we extract 2D slices and solve the acoustic wave equation \eqref{eq:acoustic} on each slice to obtain the corresponding wavefields and shot gathers.
We repeat this for five different source bandwidths (\cref{tab:bands}), motivated by multiscale FWI \citep{bunks1995multiscale,fichtner2011full}, where practitioners progressively increase the source bandwidth to mitigate cycle skipping.
For additional details on the seismic data generation process, see \cref{app:seismic_forward}.

\begin{figure}[t]
    \centering
    \includegraphics[width=\linewidth]{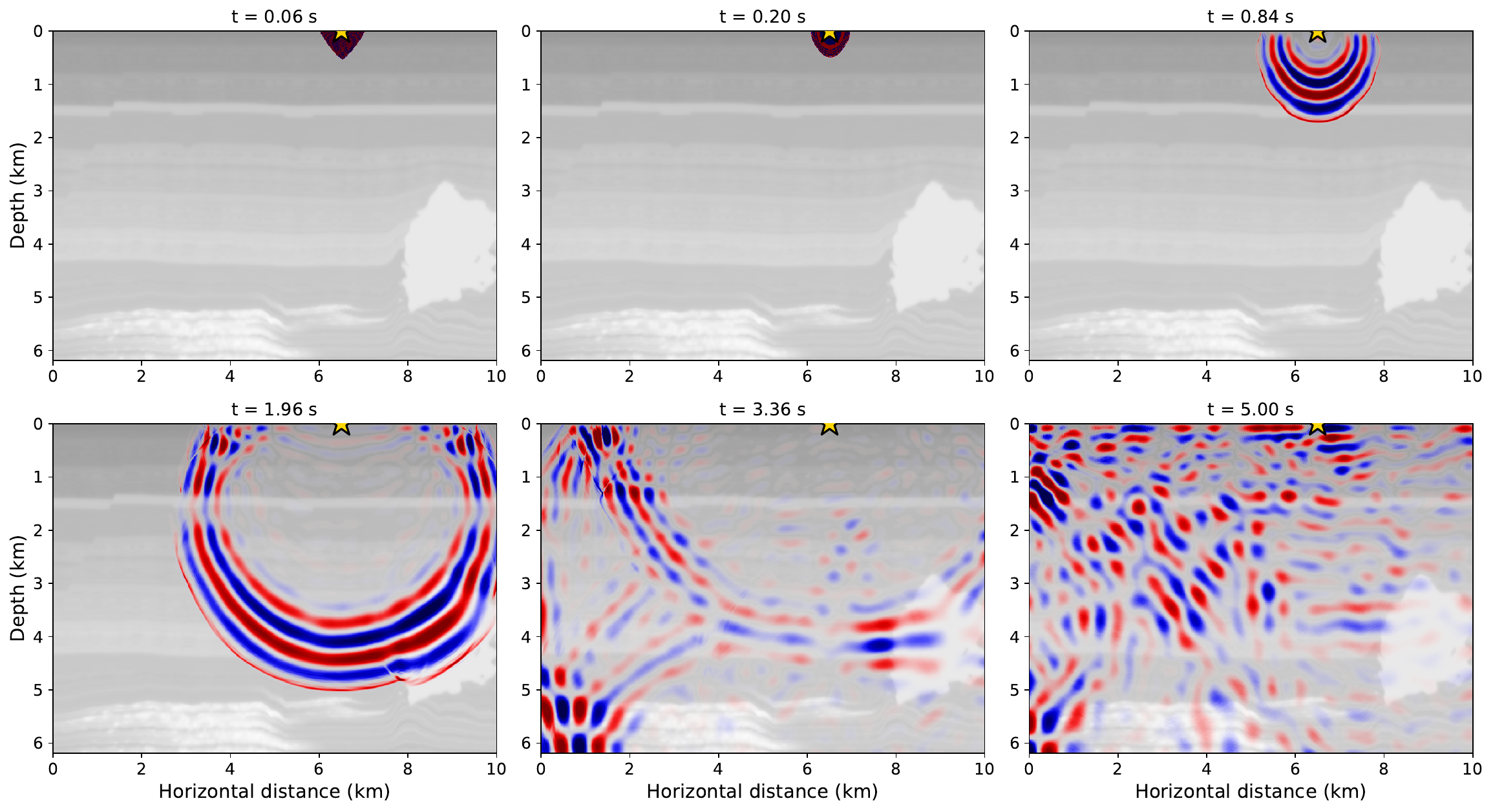}
    \caption{Wavefield evolution on a Penobscot crossline slice with a Ricker wavelet source near the surface.
        Waves radiate outward from the source until they enter a salt body (1.96 s), with chaotic multipath behavior dominating at late times (5.0 s).}
    \label{fig:wavefield_evolution}
\end{figure}

\subsection{Seismic Data Generator}

SubsurfaceGen extracts 2D slices from a 3D velocity model along a horizontal axis (inline or crossline).
SubsurfaceGen uses \devito \citep{louboutin2019devito} to solve \eqref{eq:acoustic} for each slice at a user-specified spatial discretization and time discretization, simulation time, source bandwidth, and source position.
The source $s(\mathbf{x}, t)$ is a Ricker wavelet, as motivated in \cref{sec:wave_equation}.
The user can save the wavefield $p(\mathbf{x}, t)$ and/or the shot gather $\Rc p$.

\subsection{Generation of Seismic Data for \huggingface Dataset}
\label{subsec:seismic_generation_huggingface}

\paragraph{Slice configuration.}
\label{sec:slice_extraction}
For all 42 velocity models from \cref{subsec:model_generation_huggingface}, we randomly draw 2D slices along the inline and crossline axes, with half of the slices coming from each orientation.
Each slice is of size $619 \times 1000$ at 10 m grid spacing; per-setting slice counts are listed in \cref{tab:splits}.

\paragraph{Wavefields and shot gathers.}
Our dataset targets marine acquisitions: for each slice, we solve the acoustic wave equation \eqref{eq:acoustic} with \devito on a 10 m grid, setting the source $s(\mathbf{x}, t)$ to be a Ricker wavelet near the surface to emulate an air gun towed behind a vessel.
We generate wavefields by using a single source at a random horizontal position, simulating for 5 s, and saving the wavefield $p(\mathbf{x}, t)$ as a tensor of shape $358 \times 619 \times 1000$; \cref{fig:wavefield_evolution} shows a wavefield generated from a Penobscot slice.
We generate shot gathers by using 64 equally-spaced sources and 1000 equally-spaced receivers, simulating for 8 s (long enough for the deepest reflections to return to the surface), and aggregating the individual shot gathers $\Rc p$ as a \textit{shot-gather cube} of shape $64 \times 572 \times 1000$.
Representative gathers from three training settings appear in \cref{fig:shotgather_main}.

\paragraph{Dataset splits.}
We construct two test sets (\cref{tab:splits}).
The in-distribution test set draws slices from the same five settings as training, but at non-overlapping positions within each velocity model.
The out-of-distribution test set is Penobscot, which we hold out from training entirely to evaluate generalization to unseen geology.

\begin{table}[h]
    \centering\small
    \caption{Dataset split.
        Each slice is paired with a wavefield and a shot-gather cube at every frequency band, yielding $(4{,}096 + 100 + 80) \cdot 5 = 21{,}380$ wavefields and 21{,}380 shot-gather cubes.}
    \label{tab:splits}
    \begin{tabular}{lrl}
        \toprule
        \textbf{Split}   & \textbf{\# Slices} & \textbf{Model Types}                  \\
        \midrule
        Train            & 4{,}096            & F3, GoM, Fault, Salt Canopy, SEAM     \\
        Test in-dist     & 100                & Same settings, non-overlapping slices \\
        Test out-of-dist & 80                 & Penobscot only                        \\
        \bottomrule
    \end{tabular}
\end{table}
\section{Wavefield Prediction with Neural Operators}
\label{sec:wavefield_prediction_experiments}

We train neural operators to predict wavefields---an alternative to the expensive PDE solves that dominate FWI's compute cost.
At field scale, predicting the full $T = 348$ frame trajectory on the $619 \times 1000$ grid in one forward pass exceeds GPU memory, forcing a \textit{chunked autoregressive rollout}: each chunk takes the last $t_\mathrm{in} = 10$ frames, the velocity model, and a source mask, predicts the next $t_\mathrm{out} = 50$ frames, and feeds its output into the next chunk.
This regime does not arise on smaller datasets like OpenFWI, where a single forward pass fits in memory.
We compare two architecturally distinct neural operators, TFNO \citep{kossaifi2024tfno} and DPOT \citep{hao2024dpot}, trained for 10 epochs on a single seed (full hyperparameters and architecture details in \cref{app:wavefield_prediction}).
One F3 test slice serves as a running example throughout this section.

\begin{figure}[t]
    \centering
    \includegraphics[width=\linewidth]{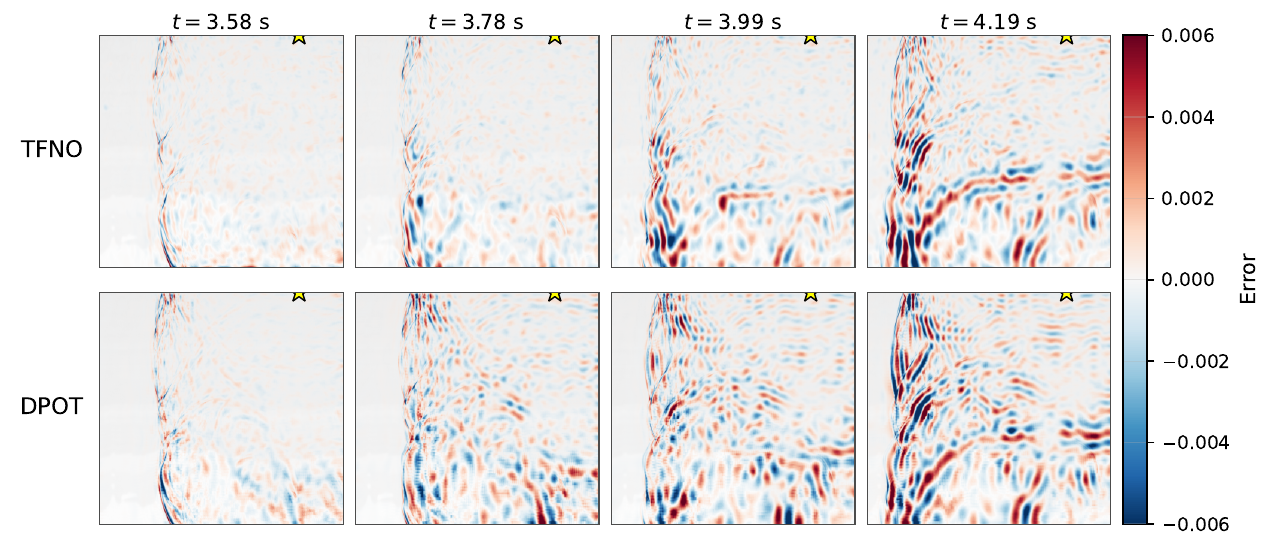}
    \caption{Wavefield prediction error for TFNO (top) and DPOT (bottom) on the F3 example slice over a single chunk at late propagation.
        Errors concentrate where the wavefield is most chaotic, illustrating the late-time scattering both networks struggle to reproduce.}
    \label{fig:chunk_grid}
\end{figure}

\cref{fig:chunk_grid} shows TFNO and DPOT prediction error on a single chunk of the F3 example slice at late propagation time, where reflections from the salt body produce chaotic multipath interference.
Both architectures track the dominant wavefronts well, with errors concentrated in deep regions.
Per-geology and per-wavenumber breakdowns of TFNO and DPOT performance are in \cref{app:wavefield_prediction:breakdowns}.

\begin{figure}[t]
    \centering
    \includegraphics[width=0.85\linewidth]{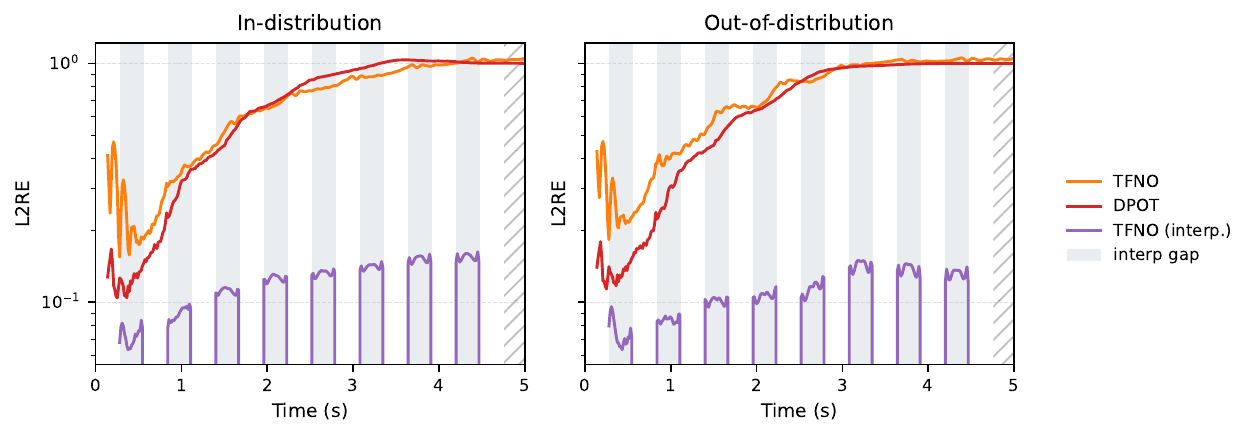}
    \caption{L2RE versus propagation time for TFNO, DPOT, and TFNO-interp on the in-distribution and out-of-distribution test sets.
        The forward operators' (TFNO, DPOT) error compounds over time; TFNO-interp's prediction windows (shaded light gray) stay roughly an order of magnitude lower.}
    \label{fig:l2_vs_time}
\end{figure}

\cref{fig:l2_vs_time} shows the cost of chunked autoregressive rollout over the entire $T = 348$ frame trajectory: the L2 relative error (L2RE) compounds by roughly an order of magnitude from one chunk to the end---though the dominant reflections remain visible in both models' predictions out to several seconds.
The field-scale dataset enables this regime in the first place: at the shorter propagation times of smaller datasets like OpenFWI (1\,s vs.\ our 5\,s), no chunking would be required, so the long-horizon failure modes we observe here would not arise.

\begin{figure}[t]
    \centering
    \includegraphics[width=\linewidth]{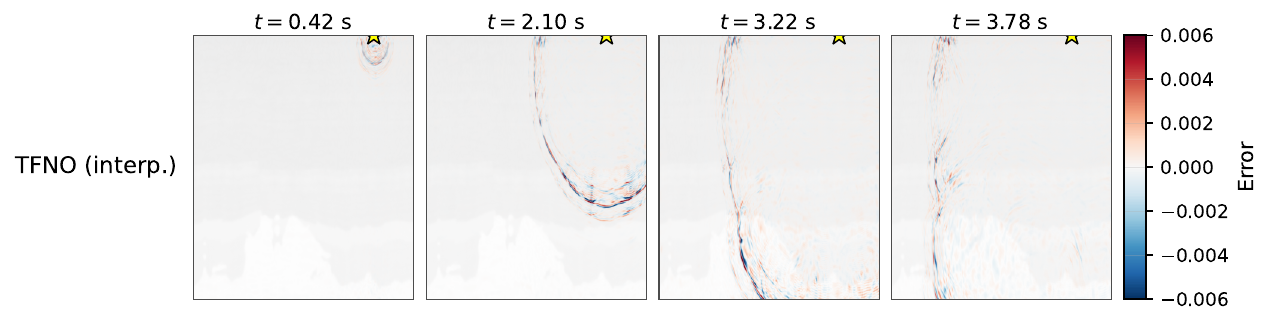}
    \caption{TFNO-interp prediction error on the same F3 example slice, sampled at four interior gap centers over the full trajectory.
        TFNO-interp's error is visibly lower than TFNO's and DPOT's, and the gap-centered sampling shows that the improvement is not just near the anchor frames.}
    \label{fig:gt_vs_interp}
\end{figure}

FWI manages a similar storage-vs-compute trade-off via optimal checkpointing \citep{symes2007checkpointing}: anchor wavefield states are stored at sparse times and the wavefield is recomputed between them by finite-difference solves.
We adapt optimal checkpointing to neural operators by filling a $\tau = 20$ frame interior gap between two stored anchor windows of $T_L = T_R = 20$ frames on each side, with the prediction pinned to the anchor frames at the gap boundaries (full architecture in \cref{app:wavefield_prediction:architectures}); we call this variant \textit{TFNO-interp}.
Anchoring at the gap boundaries uniformly reduces drift: \cref{fig:gt_vs_interp} shows it qualitatively on the F3 example, and \cref{fig:l2_vs_time} confirms TFNO-interp's L2RE in the gap intervals is roughly $6\times$ lower than either forward operator's at every time step.
These experiments show the value of the field-scale dataset: the drift on chunked rollout and the analogy to optimal checkpointing both surface at field-scale, motivating TFNO-interp.

\section{End-to-End Inversion with Encoder--Decoders}
\label{sec:inversion_experiments}

We train networks that map a 3D shot-gather cube directly to the corresponding 2D velocity model $v_p(\mathbf{x})$---an alternative to classical FWI that could avoid cycle skipping and the cost of repeated PDE solves.
To our knowledge, our dataset is the first to enable cross-geology generalization studies on this task at field-scale: Marmousi and SEAM each contain only one geological setting, and OpenFWI evaluates cross-geology generalization, but only on 0.7 km $\times$ 0.7 km models.

We compare three encoder--decoder architectures: \textbf{InversionNet} \citep{wu2020inversionnet} treats the shot-gather cube as a 2D image with 64 channels; \textbf{Transformer} \citep{liu2021swin, hatamizadeh2022swinunetr} and \textbf{CNN} \citep{isensee2024nnunetrevisited} treat it as a 3D volume.
The Transformer and CNN share an identical decoder architecture, so any gap between them is attributable to the encoder.
We train each architecture for 100 epochs across three seeds and report root mean square error (RMSE) and structural similarity (SSIM) \citep{wang2004ssim} on the three splits (\cref{tab:splits}), where Penobscot is held out as the out-of-distribution (OOD) test split.
Full details are available in \cref{app:inversion}.

\begin{figure}[t]
    \centering
    \includegraphics[width=\linewidth]{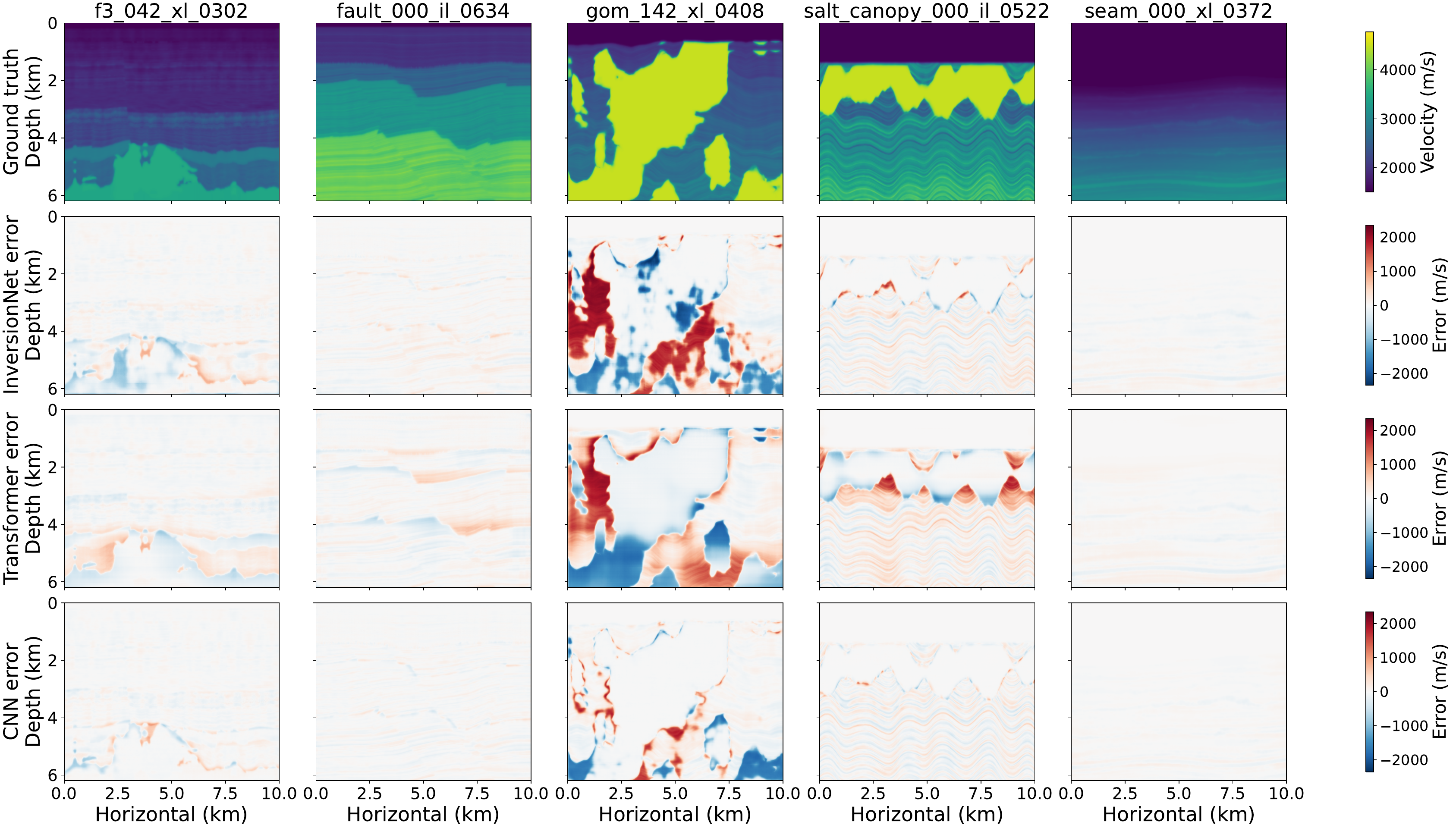}
    \caption{Inverted velocity models on the in-distribution test split (one slice per setting: F3, Fault, Gulf of Mexico, Salt Canopy, SEAM). Top row: ground-truth velocity. Subsequent rows: prediction error for InversionNet, Transformer, and CNN.}
    \label{fig:inversion_predictions_id}
\end{figure}

\cref{tab:inversion_metrics} reports RMSE and SSIM across the three splits and \cref{fig:inversion_predictions_id} shows prediction errors on the in-distribution test split.
The CNN captures large-scale geological features (salt bodies, faults) better than the other two architectures, but all three architectures struggle to resolve fine-scale features.
Moreover, the CNN uniformly outperforms InversionNet, showing the value of a 3D representation in the encoder.
The Penobscot hold-out lets us understand how each architecture generalizes to a new geological setting: the Transformer's SSIM is stable while the CNN's and InversionNet's drop by 0.04--0.07, revealing a relationship between architecture and OOD performance.

\begin{table}[t]
    \centering
    \small
    \caption{End-to-end inversion accuracy across architectures and splits (averaged over 3 seeds).}
    \label{tab:inversion_metrics}
    \begin{tabular}{l ccc ccc}
        \toprule
                     & \multicolumn{3}{c}{RMSE (m/s) $\downarrow$} & \multicolumn{3}{c}{SSIM $\uparrow$}                                                                       \\
        \cmidrule(lr){2-4} \cmidrule(lr){5-7}
        Architecture & Train                                       & Test (in-dist.)                     & Test (out-of-dist.) & Train & Test (in-dist.) & Test (out-of-dist.) \\
        \midrule
        InversionNet & 72.18                                       & 260.32                              & 603.09              & 0.909 & 0.893           & 0.848               \\
        Transformer  & 141.45                                      & 239.87                              & 526.44              & 0.881 & 0.892           & 0.883               \\
        CNN          & 55.72                                       & 135.75                              & 458.58              & 0.923 & 0.923           & 0.851               \\
        \bottomrule
    \end{tabular}
\end{table}

\section{Conclusion}
\label{sec:conclusion}

We introduce SubsurfaceGen, a GPU-accelerated procedural velocity model builder and seismic data generator for ML-based FWI.
We use SubsurfaceGen to generate an example field-scale dataset, available on \huggingface.
Our experiments reveal how SubsurfaceGen can impact ML-based FWI: for wavefield prediction, the field-scale grid forces predictions to be chunked, which motivates checkpointing strategies; for end-to-end inversion, geological diversity supports cross-geology generalization studies, opening a new possibility for evaluating architectures.
Natural extensions include wavefield and shot gather generation over 3D domains and modeling with the elastic wave equation.

\section*{Impact Statement}
SubsurfaceGen and the field-scale dataset democratize ML for seismic imaging by lowering the barrier to entry for researchers who lack proprietary seismic data.
SubsurfaceGen and the field-scale dataset can be applied for verifying the integrity of carbon sequestration sites (climate-positive) and for hydrocarbon exploration, which is a significant driver of greenhouse gas emissions (climate-negative); the net environmental impact depends on who builds upon this work and how they use it.
The field-scale dataset is intended to support ML method development and could serve as a foundation for future benchmarks, but we caution against deploying models trained on it in real-world settings without further validation and testing.
\begin{ack}
    We would like to thank Bob Clapp for creating the first version of the velocity model builder and Rustam Akhmadiev for suggesting the checkpointing idea.
    We would also like to thank Elliana Abrahams, Guillaume Barnier, Ettore Biondi, Brian Chivers, Thomas Cullison, Stuart Farris, Zachary Frangella, Wenzhi Gao, Joshua Rines, Drew Stump, Yinjun Wang, and Anders Wikum for valuable feedback that helped shape this submission.
    JS acknowledges support from the Stanford Geophysics Ph.D. program, including the secondary research project completed under the advising of CYL as part of the Ph.D. breadth requirement, and graduate research support under the advising of Biondo Biondi.
    PR and MU gratefully acknowledge support from the Office of Naval Research under award N000142412306, Air Force Office of Scientific Research under award FA9550-26-1-0012,
    the Alfred P. Sloan Foundation, the Stanford Institute for Human-Centered Artificial Intelligence, and from IBM Research as a founding member of Stanford Institute for Human-centered Artificial Intelligence.
    CYL gratefully acknowledges support from the Alfred P. Sloan Foundation via grant FG-2024-21649.
    We acknowledge Stanford University and the Center for Computation at the Stanford Doerr School of Sustainability for computational resources and support, including access to the Sherlock computing cluster and the serc partition.
    We also acknowledge computational resources provided through allocation EES260056 from the Advanced Cyberinfrastructure Coordination Ecosystem: Services \& Support (ACCESS) program, supported by the U.S. National Science Foundation.
\end{ack}

\bibliographystyle{plainnat}
\bibliography{references}


\appendix
\section{Related Work}
\label{app:related_work}

We discuss related work along three threads: datasets used to train and evaluate ML-based FWI (\cref{subsec:popular_datasets}), machine learning methods for seismic modeling and inversion (\cref{subsec:seismic_no}), and software tools for building velocity models (\cref{subsec:model_builders}).

\subsection{Datasets for ML-based FWI}
\label{subsec:popular_datasets}

Marmousi \citep{versteeg1994marmousi,martin2006marmousi2}, SEAM \citep{fehler2011seam}, OpenFWI \citep{deng2022openfwi}, $\mathbf{\mathbb{E}^{\text{FWI}}}$ \citep{feng2023efwi}, and GlobalTomo \citep{li2025globaltomo} are well-known datasets for ML-based FWI.
Marmousi, SEAM, and OpenFWI are the most closely related to SubsurfaceGen.
Each of these datasets falls short on at least one of the four properties from \cref{tab:resource_comparison}---field-scale, geologically diverse, physically realistic, extendable---in the following respects:

\begin{itemize}
    \item \textbf{Static datasets.} Existing datasets are fixed collections, which cannot support the experiments required for ML-based FWI to progress.
          Studying generalization across geological settings, scaling behavior with dataset size, and robustness to distribution shift all require the ability to procedurally generate new data with known properties.
    \item \textbf{Limited spatial extent.} The majority of OpenFWI's 2D velocity models span only 0.7 km $\times$ 0.7 km.
          This is problematic because cycle skipping, illumination gaps, and low-frequency recovery all worsen as spatial extent grows: methods evaluated only at small scales provide limited evidence of their performance on real-world surveys.
    \item \textbf{Limited temporal extent.} The majority of OpenFWI's 2D shot gathers span only 1 s, which limits the maximum imaging depth to the upper 1–2 km of the subsurface, falling well short of the depths where carbon storage sites and hydrocarbons are often found.
    \item \textbf{Limited geological diversity.} Marmousi captures complex folding and faulting, and SEAM captures salt bodies, but neither contains clinoforms, carbonate platforms, or velocity interbedding.
          OpenFWI's 2D families lack salt bodies, clinoforms, and velocity interbedding; its 3D family achieves realistic scale (4 km $\times$ 4 km $\times$ 3.5 km, 5 s of recording) but represents a single geological scenario and therefore inherits the same diversity limitations.
          These omissions correspond to the hardest and most application-relevant regimes of FWI: salt bodies produce strong contrasts that are challenging to invert, carbonate platforms and clinoforms define the sedimentary settings targeted for carbon sequestration and hydrocarbon exploration, and velocity interbedding---thin alternating layers that are difficult to resolve---determines caprock integrity and reservoir quality.
    \item \textbf{Lack of physical realism.} Many of OpenFWI's 2D velocity models contain visible artifacts that make them physically unrealistic; models trained on them risk learning to reproduce these artifacts rather than geologically meaningful structure.
\end{itemize}

$\mathbf{\mathbb{E}^{\text{FWI}}}$ extends OpenFWI to the elastic wave equation, but inherits OpenFWI's limited spatial/temporal extent and limited geological diversity.
GlobalTomo is intended for whole-Earth seismology (mantle structure, earthquake source mechanics, planetary tomography) rather than the sedimentary settings targeted by SubsurfaceGen.

SubsurfaceGen addresses all of the limitations above, since it enables procedural generation of field-scale, geologically diverse, physically realistic 3D velocity models and seismic data.

\subsection{Machine Learning for Seismic Modeling and Inversion}
\label{subsec:seismic_no}

Wavefield prediction with neural operators \citep{yang2021seismic, yang2023unoseismic, zhang2023elasticfno, huang2025scattered} has been restricted to small domains (less than one square kilometer) with unrealistic geology (random textures or simple shapes instead of layered, faulted, or salt-containing models) and acquisition geometries unlike real surface surveys (e.g., putting sources on the entire perimeter of the model).
SubsurfaceGen provides the data needed to train and evaluate these operators in production-relevant settings, as shown in \cref{sec:wavefield_prediction_experiments}.
A second line of work learns the inverse map from shot gathers directly to velocity models using neural networks \citep{arayapolo2018dltomography, yang2019fcnvmb, wu2020inversionnet, zhang2020velocitygan, wang2023svit}.
\citet{jin2024bigfwi} scales this approach to 408{,}000 samples, but is restricted to the small spatial extent of OpenFWI (0.7 km).
Only \citet{farris2023learning, farris2023thesis} has trained end-to-end inversion at field scale---on $\sim$17\,km $\times$ 12\,km models in the Gulf of Mexico, using a data engine that was never publicly released.
SubsurfaceGen (and the included field-scale dataset) fills this void, and we train inversion baselines on it in \cref{sec:inversion_experiments}.
A third line of work uses generative models trained on velocity models as data-driven priors for FWI \citep{mosser2020ganprior, stitt2023deepdix, wang2023priorfwi, stitt2025latentdiff}; SubsurfaceGen provides a distribution of geologically plausible velocity models required for training these priors.

\subsection{Model Building Tools}
\label{subsec:model_builders}

Structural geomodeling tools like pynoddy \citep{wellmann2016pynoddy}, GemPy \citep{delavarga2019gempy}, and LoopStructural \citep{grose2021loopstructural} are unable to encode the pixel-level velocity heterogeneity (fractal noise, fine interbedding) that defines the sedimentary basin settings SubsurfaceGen targets.
Moreover, they lack a pipeline to generate training data pairs of velocity models and seismic data.
The closest peer to SubsurfaceGen is Synthoseis \citep{merrifield2022synthoseis}, a CPU-based pipeline that produces seismic data using 1D reflectivity convolution rather than full wave physics, so the resulting traces lack important phenomena like multipathing and chaotic late-time interference.
Synthoseis also omits modules for realistic salt geometries, carbonate platforms, and clinoforms.
SubsurfaceGen closes these gaps: it is GPU-enabled, contains salt, carbonate platform, and clinoform modules, and produces seismic data by solving the acoustic wave equation.
\section{Velocity Model Builder: Modules}
\label{app:model_builder_modules}

This appendix gives algorithmic details for the eight builder modules
introduced in \cref{sec:model_builder} (summarized in
\cref{tab:module_summary}).
Each module is described in its own subsection
(\cref{app:model_builder_modules:deposit,app:model_builder_modules:squish,app:model_builder_modules:fault,app:model_builder_modules:saltsdt,app:model_builder_modules:saltwedge,app:model_builder_modules:carbonate,app:model_builder_modules:delta,app:model_builder_modules:sos}).
The companion appendices show how these modules are composed into each
geological setting (\cref{app:model_details}) and a CPU vs. GPU timing
benchmark of the model builder (\cref{app:builder_benchmark}).
\textbf{This section leans heavily on geological terminology.}

\begin{table}[h]
    \small
    \centering
    \caption{Catalog of modules in the SubsurfaceGen model builder.
        Each row describes a module using standard geological terminology.}
    \label{tab:module_summary}
    \begin{tabular}{l p{0.7\linewidth}}
        \toprule
        \textbf{Module}        & \textbf{What it emulates}                                                                                                                        \\
        \midrule
        \deposit               & A sediment package laid down between two bedding surfaces, with thickness, base $V_p$, depth gradient, and interbed reflectivity texture inside. \\
        \addlinespace
        \squish                & Basin-scale vertical warping of all layers above the operator (anticlines, broad post-rift sag) along a single azimuth.                          \\
        \addlinespace
        \fault                 & Applies a localized coordinate warp across a curved fault surface, creating layer offsets and fault-related displacement.                        \\
        \addlinespace
        \saltsdt               & Inserts one or more irregular high-velocity salt bodies, generated from perturbed 3D ellipsoidal masks.                                          \\
        \addlinespace
        \saltwedge             & The deformation of nearby sediment layers against the flanks of a salt body: layers curving upward against the flank rather than onlapping flat. \\
        \addlinespace
        \carbonateplatform     & A buried carbonate reef rim with marine-flank, lagoon, and chaotic-core facies, asymmetric along strike.                                         \\
        \addlinespace
        \deltaclinoformdeposit & Stacks curved clinoform-like bed surfaces to create shelf-to-basin layered geometries with thin internal beds.                                   \\
        \addlinespace
        \structuralsmoother    & Structure-oriented smoothing: anisotropic along bed orientation, weak across bed contacts and fault offsets \citep{hale2009structureoriented}.   \\
        \bottomrule
    \end{tabular}
\end{table}

\paragraph{Notation.}
$V$ denotes the velocity field $v_p(\mathbf{x})$ in m/s on a $(N_z, N_y, N_x)$ grid; $L$ is an integer-valued volume of the same shape recording which geologic layer each voxel belongs to. The pseudocode below shows updates to $V$ explicitly; updates to $L$ happen alongside $V$ in the implementation but are omitted for clarity. SubsurfaceGen bundles $V$ and $L$ in a single data structure.

\subsection{\deposit}
\label{app:model_builder_modules:deposit}

The \deposit module (\cref{alg:deposit}) adds a sedimentary package with interbeds
(thin layers of contrasting velocity) on top of a model.
Interbed boundaries are 2D surfaces sampled from a 3D simplex noise field (a smooth,
spatially correlated random field); each bed is assigned a single Gaussian-drawn
velocity with a global depth gradient added across the slab, and an additional 3D
simplex noise field provides intra-bed spatial texture (with a per-bed random
$z$-offset into the shared field decorrelating textures across beds).
Layer assignment is vectorized via \texttt{torch.searchsorted}.
An optional sinusoidal bedform mode (\texttt{interbed\_mode = "sinusoidal"}) bypasses
the boundary-based path and generates the slab as a sum of two sinusoidal harmonics
evaluated on a $z$-warped axis (depth axis non-uniformly stretched so bed thickness
varies; built as the cumsum of a smoothed instantaneous frequency) with added
rugosity (small-amplitude roughness), plus an additive 3D patchy-noise texture.
Unit convention: the user-facing config specifies $\bar{h}$ and $\sigma_h$ in meters;
internally these are divided by the vertical discretization $\Delta z$ to yield the
per-cell quantities used in the formulas below.

\begin{algorithm}[h]
    \caption{\texttt{Deposit.apply} (sedimentary package with interbeds)}
    \label{alg:deposit}
    \begin{algorithmic}[1]
        \Require Thickness $t$, base velocity $v_0$, depth gradient $\beta$, mean interbed thickness $\bar h$, interbed-thickness std $\sigma_h$, per-bed velocity std $\sigma_p$, boundary-noise amplitude $a_B$, texture amplitude $a_\eta$, taper $\tau(z) = \min(z/(f_\text{top} t), 1) \cdot \min((t-z)/(f_\text{base} t), 1)$.
        \State Subdivide the slab into $n \approx \lfloor t/\bar h \rfloor$ thin beds: draw thicknesses $\sim \mathcal{N}(\bar h, \sigma_h^2)$ (clamped to $\ge \tfrac{1}{2}$ sample), cumsum into internal boundary depths $\{H_i\}_{i=1}^{n-1}$.
        \State $H_i(y, x) \gets H_i + \tau(H_i)\, a_B\, \eta^{(i)}(y, x)$, clamped to $[0, t-1]$.
        \State Draw per-bed velocities $v_1, \ldots, v_n \sim \mathcal{N}(v_0, \sigma_p^2)$; let $\bar v = \tfrac{1}{n}\sum_\ell v_\ell$.
        \For{each voxel $(z, y, x)$ in the deposit band}
        \State $\ell \gets \textsc{searchsorted}(\{H_i(y, x)\}, z)$; $S_\ell$ is bed $\ell$'s start depth.
        \State $V[z, y, x] = (\bar v + \beta z) + \tau(z) \big[(v_\ell - \bar v) + a_\eta\, \eta_\text{3D}(z - S_\ell, y, x)\big]$.
        \EndFor
        \State Per-column sinc-resample $V$ vertically to conform to the underlying topography $h(y, x)$.
    \end{algorithmic}
\end{algorithm}

\subsection{\squish}
\label{app:model_builder_modules:squish}

The \squish module (\cref{alg:squish}) applies a vertical displacement to a model, emulating basin-scale warping and long-wavelength folding
without introducing faulting behavior.
The displacement is a 2D fractal noise field with configurable maximum shift, spatial wavelengths, azimuth, and octave count, which is applied to every voxel in
the column via sinc resampling.

\begin{algorithm}[h]
    \caption{\texttt{Squish.apply} (vertical warping)}
    \label{alg:squish}
    \begin{algorithmic}[1]
        \Require Maximum shift $\Delta z_\text{max}$, spatial wavelengths $(\lambda_y, \lambda_x)$, azimuth $\alpha$, octave count $n_\text{oct}$.
        \State Sample a 2D fractal-noise displacement field $\Delta z(y, x)$ with amplitude $\Delta z_\text{max}$, wavelengths $(\lambda_y, \lambda_x)$ rotated by $\alpha$, summed over $n_\text{oct}$ octaves.
        \State Pull-sample $V$ at $(z + \Delta z(y, x), y, x)$ via sinc interpolation.
    \end{algorithmic}
\end{algorithm}

\subsection{\fault}
\label{app:model_builder_modules:fault}

The \fault module (\cref{alg:fault}) applies listric normal-fault deformation to a model. It does this through building a
smooth 3D displacement field and then warping the model. Faults are parameterized by a center, azimuth, dip angle, curvature radius,
slip magnitude, direction, and die-off extents. Each voxel in the post-fault model is pull-sampled from its pre-slip position. The fault
geometry is represented as a cylindrical coordinate system where a large cylinder radius gives a planar fault while a small radius creates
a curved listric fault (l-shaped fault).

\cref{alg:fault} represents fault slip or displacement as a rotation around this cylinder, with cosine tapers controlling
how the displacement dies off across the fault, along strike, and near the ends of the rupture. For each output voxel, the code computes where
that voxel came from in the original unfaulted model, then samples the original model at that source location. Continuous fields such as velocity
are resampled with sinc interpolation, while integer fields such as layer labels are resampled with nearest-neighbor interpolation.
The implementation also supports rough fault surfaces by perturbing the fault radius with fractal noise, and can generate smaller subsidiary
faults around a primary fault using simplified Riedel-shear geometry.

\begin{algorithm}[h]
    \caption{\texttt{Fault.apply} (listric normal-fault deformation, single fault)}
    \label{alg:fault}
    \begin{algorithmic}[1]
        \Require Fault center $c$, strike azimuth $\alpha$, dip $\theta$, curvature radius $R$, slip $s$, direction $d \in \{+1, -1\}$, die-off extents $(d_\text{die}, d_\text{perp}, \theta_\text{die})$.
        \State Cylinder center $\tilde c = c - (\delta_\text{az}\sin\alpha,\ \delta_\text{az}\cos\alpha,\ \delta_z)$ with $(\delta_z, \delta_\text{az}) = (R\sin\theta, R\cos\theta)$; reference angle $\theta_0 = \operatorname{atan2}(\delta_z, \delta_\text{az})$.
        \State Slip as rotation: $\theta_\text{shift} = s/(2\pi R) \cdot 360^\circ$.
        \For{each voxel $(z, y, x)$}
        \State Cylindrical coordinates of $(z, y, x) - \tilde c$ in the fault-aligned frame: radius $r$, angle $\theta_\text{old}$, along-strike offset $\text{az}_\text{strike}$.
        \State Die-off ratios $\rho_r = |r - R|/d_\text{die}$, $\rho_\theta = |\theta_\text{old} - \theta_0|/\theta_\text{die}$, $\rho_s = |\text{az}_\text{strike}|/d_\text{perp}$.
        \State Taper $w(x) = \cos(\tfrac{\pi}{2}\rho_r)\cos(\tfrac{\pi}{2}\rho_\theta)\cos(\tfrac{\pi}{2}\rho_s) \mathbf{1}\{\rho_r, \rho_\theta, \rho_s < 1\}$.
        \State Side $\chi = +1$ if $r \geq R$ (footwall), else $\chi = -1$ (hanging wall); $\theta_\text{new} = \theta_\text{old} - \chi\, d\, w(x)\, \theta_\text{shift}$.
        \State Forward-mapped position $p_\text{fwd}(z, y, x) \gets$ inverse cylindrical transform of $(r, \theta_\text{new}, \text{az}_\text{strike})$.
        \State Pull source $\text{src}(z, y, x) = 2(z, y, x) - p_\text{fwd}(z, y, x)$ (Cartesian-linearized inverse, exact to first order in $\theta_\text{shift}$).
        \EndFor
        \State Pull-sample $V$ at $\text{src}$ via sinc; voxels with $\text{src}_z \notin [0, N_z)$ get fill values.
    \end{algorithmic}
\end{algorithm}

\subsection{\saltsdt}
\label{app:model_builder_modules:saltsdt}

The \saltsdt module (\cref{alg:saltsdt}) generates irregular 3D salt bodies, similar to high-velocity salt
structures in the Gulf of Mexico using a signed distance transform perturbed by a Gaussian random field (GRF).
Each body starts from a randomly placed and rotated ellipsoid, which provides a
simple controllable shape for the salt core. The
ellipsoid mask is converted into a signed distance field, so the salt boundary
can be modified smoothly by adding a spatially correlated Gaussian random field.
The correlation lengths control the preferred scale and direction of boundary
variations, while the perturbation amplitude controls how far the final body
deviates from the original ellipsoid. Smaller spectral exponents preserve more
short-wavelength roughness, whereas larger values produce smoother salt
boundaries. After all bodies are generated, their masks are merged, cleaned with
morphological opening and closing, and small isolated components are removed.
The final mask is prevented from overwriting water and is then assigned the
configured salt velocity.

\begin{algorithm}[h]
    \caption{\texttt{SaltSDT.apply} (signed distance transform salt bodies)}
    \label{alg:saltsdt}
    \begin{algorithmic}[1]
        \Require Body count $n_b$, placement bounds, base radius range, anisotropic correlation length fractions $(f_{\ell_x}, f_{\ell_y}, f_{\ell_z})$, GRF amplitude $a$, spectral exponent $p$, salt velocity $v_s$, water threshold $v_\text{water}$.
        \State $K \gets \mathbf{0}$.
        \For{$i = 1, \ldots, n_b$}
        \State Sample center, per-axis radii $(r_x, r_y, r_z)$, rotation angles; let $\bar r = (r_x + r_y + r_z)/3$.
        \State Rasterize rotated ellipsoid; compute signed distance $\phi^{(i)}$ via Meijster Euclidean distance transform.
        \State Sample GRF $\xi^{(i)}$ from power spectrum $P(\mathbf{k}) \propto \big(1 + \sum_k (k_k \ell_k)^2\big)^{-p}$ with $\ell_k = \max(r_k f_{\ell_k}, 5)$.
        \State $\phi'^{(i)}(x) = \phi^{(i)}(x) + a\, \bar r\, \xi^{(i)}(x)$; $K_i = \mathbf{1}\{\phi'^{(i)} > 0\}$; $K \gets K \lor K_i$.
        \EndFor
        \State Morphologically open then close $K$; drop connected components smaller than 1000 voxels.
        \State $K \gets K \land (V \geq v_\text{water})$; $V[K] \gets v_s$.
    \end{algorithmic}
\end{algorithm}

The geometric structure of the salt bodies is controlled by two sets of parameters:
$(\ell_x, \ell_y, \ell_z)$ (anisotropy of the boundary correlation length)
and $(a, p)$ (amplitude and spectral falloff of the perturbation):
increasing $\ell_z$ relative to $\ell_x$ elongates bodies vertically (as
in the Penobscot edge diapir); decreasing $p$ raises high-frequency
roughness on the body boundary; increasing $a$ pushes bodies further from
their base ellipsoid.

\subsection{\saltwedge}
\label{app:model_builder_modules:saltwedge}

The \saltwedge module (\cref{alg:saltwedge}) deforms the sediment column in the neighborhood
of salt bodies (produced by \saltsdt) to generate flanking onlap and
drape geometries consistent with the halokinetic sequence framework of
\citet{giles2012halokinetic}. The goal is not to simulate salt mechanics directly,
but to create the common seismic appearance of sediment layers bending, draping,
or onlapping near salt bodies. The standard mode computes larger upward shifts
near salt and smaller shifts farther away. This shift is strongest near steep
salt flanks, decays with distance from the salt, can taper with depth, and is
modulated by smooth fractal noise so the deformation is not perfectly symmetric.
An optional near-salt squeeze component compresses layers close to the salt
flank when its strength is set above zero (off by default).
An optional top-conforming mode separately detects the top of the salt body and
pushes overlying sediments upward above salt crests, producing diapir-like drape.
After the drag, optional squeeze, and optional top-conforming components are
combined, the displacement field is clamped to nonnegative values, optionally
smoothed in the horizontal plane, made monotone with depth (enabled by default),
and then applied as a vertical-only pull warp.
Salt voxels themselves are preserved at the configured salt velocity.

\begin{algorithm}[h]
    \caption{\texttt{SaltWedge.apply} (sediment deformation around salt bodies)}
    \label{alg:saltwedge}
    \begin{algorithmic}[1]
        \Require Salt mask $K$, base influence $R_0$, flank-influence $R_\text{flank}$, drag exponent $p$, vertical-flank threshold $\tau$, depth taper $g(z)$ on $(z_\text{top}, z_\text{bot}, g_\text{pow})$, base amplitude $A_\text{frac}$, conform flag and parameters $(R_\text{conform}, c_\text{uplift}, s_\text{conf})$.
        \State Signed distance $\phi(x) = d_\text{out}(x; K) - d_\text{in}(x; K)$; vertical normal $\hat n_z = \partial_z \phi / \|\nabla\phi\|$.
        \State Vertical-flank set $V_\text{flank} = \{x \in \partial K : |\hat n_z(x)| < \tau\}$; $d_\text{vert}(x) = $ EDT distance to $V_\text{flank}$.
        \State Drag weights $w_d(x) = \big(\max(0, 1 - \phi/R_\text{field}(z))\big)^p$, $w_v(x) = \exp(-(d_\text{vert}/R_\text{flank})^2)$, $w = w_d w_v \mathbf{1}\{x \notin K\}$, with $R_\text{field}(z) = R_0(1 + \beta_\text{rad}\sin(\pi z/N_z))$.
        \State Drag displacement $\Delta z_\text{drag}(x) = A_\text{frac}\, N_z\, w(x)\, g(z)\, A(x)$, with $A(x)$ a fractal amplitude field.
        \If{conform enabled}
        \State Salt-top $\zeta_\text{top}(y, x) = \min\{z: K[z, y, x] = 1\}$; crest relief $R_\text{crest} = \max(0, \mathrm{Gauss}_\sigma(\zeta_\text{top}) - \zeta_\text{top})$, normalized to $[0,1]$.
        \State $\Delta z_\text{conf}(x) = c_\text{uplift} N_z s_\text{conf}\, R_\text{crest}(y, x)\, e^{-(\zeta_\text{top} - z)/R_\text{conform}}\, g(z) \mathbf{1}\{z < \zeta_\text{top}, x \notin K\}$.
        \Else
        \State $\Delta z_\text{conf} \gets 0$.
        \EndIf
        \State $\Delta z(x) = \max(0, \Delta z_\text{drag} + \Delta z_\text{conf})$; enforce monotonicity $\Delta z[z, y, x] \gets \max_{z' \leq z} \Delta z[z', y, x]$.
        \State Warp $V$ by pull-sampling at $(z + \Delta z, y, x)$ with $z$-linear interpolation; salt voxels held at $v_s$.
    \end{algorithmic}
\end{algorithm}

\subsection{\carbonateplatform}
\label{app:model_builder_modules:carbonate}

The \carbonateplatform module (\cref{alg:carbonate}) builds asymmetric reef bodies with a
steep marine flank, a gentle leeward ramp, and a ``chaotic'' core. Multiple
platforms along a shared strike direction are merged via a smooth-minimum
signed distance operator with a configurable blending parameter. Facies
velocities (core, marine forereef, lagoon, carbonate basin, sequence
boundary) are assigned separately. A lateral fill texture with interbedded
velocities and a patchy 3D noise field fills the surrounding sediment
bounded by the platform contacts.

\begin{algorithm}[h]
    \caption{\texttt{CarbonatePlatform.apply} (asymmetric reef body with facies zoning)}
    \label{alg:carbonate}
    \begin{algorithmic}[1]
        \Require Thickness $t_\text{use}$, platform azimuth $\alpha$, center count $N_p$, width fraction $f_w$, shape exponent $\kappa$, marine squash $\mu$, leeward stretch $\lambda$, growth phases $n_\text{phases}$, smooth-min parameter $k$, facies velocities, lateral-fill fraction $f_\text{fill}$.
        \State Domain-warp the horizontal grid: $(y_w, x_w) = (y, x) + a_w(\eta_y, \eta_x)$.
        \State Sample $N_p$ platform-center paths $(c_y^{(j)}(t), c_x^{(j)}(t))$ as oscillation + smoothed walk + drift along $\alpha$.
        \State For each platform $j$: rotate into marine-strike frame, split marine vs leeward by $\mu, \lambda$, and form normalized distance $\text{dist}_j(y, x) = \sqrt{(d_\text{mar}^\text{asym})^2 + d_\text{str}^2}/r_\text{eff}^{(j)}$.
        \State Blend platforms via smooth-minimum: $\text{dist}_\text{blend} = \operatorname{smin}_k(\text{dist}_1, \ldots, \text{dist}_{N_p})$.
        \State Build 2D thickness map $T(y, x)$ from $n_\text{phases}$ growth lenses $\exp(-\text{dist}_\text{blend}^\kappa)$, normalized to $[\text{min\_frac}, 1]\,t_\text{use}$.
        \State Build unmorphed volume of shape $m_\text{mound} \times N_y \times N_x$: assign facies velocity from $\{v_\text{core}, v_\text{fore}, v_\text{lag}, v_\text{basin}\}$ via $(\text{dist}_\text{blend}, \text{bias})$ where $\text{bias}_j = \tfrac{1}{2}(d_\text{mar}/|d| + 1)$ marks each voxel marine-vs-leeward; add $n_\text{ib}$ interbeds and core texture.
        \State Per-column linearly resample to $T(y, x)$ samples and place into $z \in [h - T, h)$.
        \State Off-platform columns in $[h - f_\text{fill}\,t_\text{use}, h)$ get a low-velocity interbed fill.
    \end{algorithmic}
\end{algorithm}

\subsection{\deltaclinoformdeposit}
\label{app:model_builder_modules:delta}

The \deltaclinoformdeposit module (\cref{alg:delta}) generates forestepping deltaic
packages as a stack of flat $N_c$ clinothem envelopes, each containing $B$ internal beds.
Internal beds are then created by interpolating between the
previous surface and the clinothem envelope, producing thin layers that follow
the same folded-over clinoform geometry. The module can optionally vary the
rollover position, extent, truncation, and onlap taper from one clinothem to the
next using a prescribed sequence of systems-tract labels, but these labels act
only as geometric controls on the stacked surfaces rather than as a full
process-based depositional model.

\begin{algorithm}[h]
    \caption{\texttt{DeltaClinoformDeposit.apply} (forestepping clinothem package)}
    \label{alg:delta}
    \begin{algorithmic}[1]
        \Require Basinward azimuth $\alpha$, clinothem count $N_c$, beds per clinothem $B$, envelope sharpness $(s_\text{fore}, s_\text{toe})$, flat-topset fraction $\phi_\text{flat}$, toe parameters $(u_{t,0}, d_\text{toe}, \beta_\text{toe})$, rollover advance $\Delta u_r$, optional systems-tract sequence, lobe shape, facies velocities, truncation period $n_\text{trunc}$. Let $\sigma(x) = 1/(1 + e^{-x})$; rotated $(u, v)$ coordinates with $u \in [0, 1]$ landward$\to$basinward.
        \For{$k = 1, \ldots, N_c$}
        \State Update rollover $u_{r,k} = u_{r,0} + R_k$ with cumulative advance $R_k = R_{k-1} + \Delta r_k$ (per-tract sign).
        \State Lobe taper $\Lambda(y, x)$: super-Gaussian in across-strike coordinate $v$.
        \State Envelope (two-stage composition): main foreset $z_\text{main}(u) = z_k^\text{top} + (z_k^\text{fore} - z_k^\text{top})\,\sigma(s_\text{fore}(u_\text{eff} - u'_{r,k}))$; toe stage $z_\text{env}(u) = z_\text{main} + (z_\text{toe}^* - z_\text{main})\,\sigma(s_\text{toe}(u - u_{t,0}))\,\beta_\text{toe}$.
        \For{$b = 1, \ldots, B$}
        \State Bed surface $z_{k,b} = z_{k,\text{base}} + (z_\text{env} - z_{k,\text{base}})(b/B)\,\Pi(u)\,\Lambda(y,x)$, where $z_{k,\text{base}}$ is the base of clinothem $k$ (the top of clinothem $k{-}1$) and $\Pi(u)$ is a product of $\sigma$-gated onlap, extent, and shift tapers.
        \State Bed velocity $v_b = v_\text{base} + v_\text{range}(f - 0.5)\cdot 2 + a_\text{lam}(-1)^{kB+b} + \gamma z_{k,b}$, where $f$ is the $\sigma$-blended sand fraction across topset/foreset/toe zones.
        \EndFor
        \If{$k \bmod n_\text{trunc} = 0$}
        \State Erode the top $\min(B, k)$ bed surfaces toward the stack surface (toplap sequence boundary).
        \EndIf
        \EndFor
        \State Rasterize via soft-Heaviside anti-aliasing over all $N_c B$ bed surfaces.
    \end{algorithmic}
\end{algorithm}

\subsection{\structuralsmoother}
\label{app:model_builder_modules:sos}

The \structuralsmoother module (\cref{alg:sos}) implements structure-oriented smoothing
\citep{hale2009structureoriented}.
The goal is to smooth the velocity volume
anisotropically: strongly along the local bed direction, weakly across it.
The operator estimates local orientation from a structure tensor built from
Sobel-filtered gradients of $V$, then computes an eigendecomposition of the
smoothed structure tensor to construct a per-voxel diffusion tensor.
The smoothed output is the solution of a symmetric positive-definite linear
system solved by conjugate gradient.

\begin{algorithm}[h]
    \caption{\texttt{StructuralSmoother.apply} (Structure-oriented smoothing)}
    \label{alg:sos}
    \begin{algorithmic}[1]
        \Require Tensor-smoothing scale $\sigma_\text{tensor}$, multi-scale window $W$, smoothing scale $c$, cross-layer diffusivity $\alpha \in (0, 1]$, conjugate gradient iteration budget $n_\text{cg}$.
        \State Structure tensor $T(x) = \sum_{s=1}^{W} \tfrac{1}{s}\, g_s(x)\, g_s(x)^\top$ with $g_s = \nabla V$ at scale $s$; Gaussian-smooth each component at scale $\sigma_\text{tensor}$.
        \State Eigendecompose $T$ pointwise into $\lambda_u \geq \lambda_v \geq \lambda_w$ with eigenvectors $u, v, w$; build diffusion tensor $D(x) = \alpha\, uu^\top + vv^\top + ww^\top$.
        \State Solve $(I + c\, G^\top D G)\, y = V$ by $n_\text{cg}$ conjugate gradient iterations, where $G$ is the discrete 3D gradient operator.
    \end{algorithmic}
\end{algorithm}

\section{Velocity Model Builder: Model Generation for Field-Scale Dataset}
\label{app:model_details}

This appendix describes the model building process for the six geological settings in our field-scale dataset on \huggingface.
For each setting we give (i) a description of the model, (ii) the real-world geological settings that inspired it (with supporting references), (iii) the
module builder sequence with module configurations, and (iv) what geological features we deliberately leave out.
The modules are in \cref{app:model_builder_modules}; CPU vs. GPU
timings for the model builder are in \cref{app:builder_benchmark}.
\textbf{This section leans heavily on geological terminology.}

\begin{figure}[t]
    \centering
    \includegraphics[width=\linewidth]{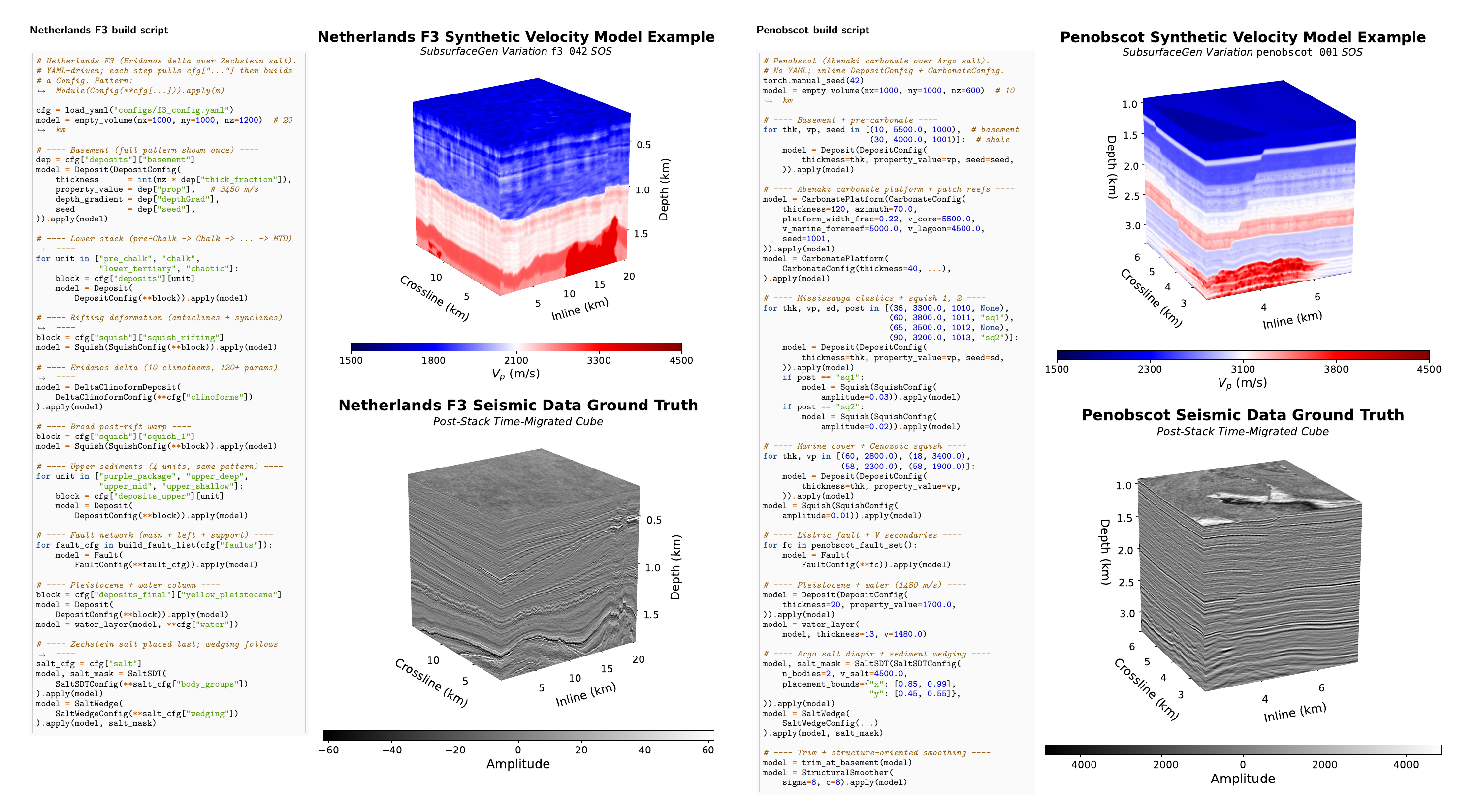}
    \caption{Python build scripts and (velocity model, migrated cube) pairs for Penobscot (left) and F3 (right).
        Each build script is a sequence of \texttt{Module(Config(\dots)).apply(model)} calls that create geological features in the velocity model.
        Each call corresponds to a feature visible in the migrated cube, e.g., the carbonate platform in Penobscot and the clinothem stack and salt diapir in F3.}
    \label{fig:build_panels}
\end{figure}

Every velocity model has shape $619 \times 1000 \times 1000$ at 10 m
isotropic spacing.
\cref{tab:model_inventory} gives the full inventory and
\cref{fig:build_panels} provides build scripts and visualizations of models and migrated cubes for the Penobscot and F3 settings.

\begin{table}[h]
    \centering
    \small
    \caption{Velocity model inventory.
        The velocity models all have identical shape and are at 10 m resolution.
        Penobscot is held out as the out-of-distribution test set during training; the
        remaining 41 models populate training and in-distribution test splits.}
    \label{tab:model_inventory}
    \begin{tabular}{llll}
        \toprule
        \textbf{Name}   & \textbf{Volumes} & \textbf{Shipped shape}        & \textbf{Setting}                              \\
        \midrule
        F3              & 10               & $619 \times 1000 \times 1000$ & Dutch North Sea, deltaic shelf with deep salt \\
        GoM             & 10               & $619 \times 1000 \times 1000$ & Gulf of Mexico, allochthonous salt bodies     \\
        Fault           & 5                & $619 \times 1000 \times 1000$ & Dense normal-fault basin (stress test)        \\
        Salt Canopy     & 4                & $619 \times 1000 \times 1000$ & Gulf of Mexico, tabular salt canopy           \\
        SEAM (Phase I)  & 12               & $619 \times 1000 \times 1000$ & Deepwater GoM, reused industry model          \\
        Penobscot (OOD) & 1                & $619 \times 1000 \times 1000$ & Scotian Shelf, carbonate platform             \\
        \bottomrule
    \end{tabular}
\end{table}

\subsection{Penobscot}
\label{app:model_details:penobscot}

Our Penobscot model is a thick carbonate platform at depth, overlain by a
siliciclastic column with a main listric fault and a small salt diapir at
one edge.
The setting is the LaHave Platform on the Scotian Shelf (offshore
Nova Scotia): a Mesozoic passive margin that accumulated kilometers of
post-rift sediment over a Late Jurassic carbonate bank
\citep{wade1990geology, brown2008regional}.
The Abenaki Formation is the
geologic anchor \citep{eliuk1978abenaki, weissenberger2006panuke}.
The Penobscot 3D survey itself is contributed by
CNSOPB and dGB Earth Sciences \citep{baroni2019penobscot}.
The field has no
thick allochthonous salt the way the Gulf of Mexico does; its salt
contribution is subordinate and mostly confined to the adjacent Sable
Subbasin \citep{kidston2002hydrocarbon, brown2008regional}.
We create a minor diapir in the synthetic to add some halokinetic texture in an
otherwise salt-poor geology.

The builder runs a deterministic
21 step sequence on a $1000 \times 1000$ grid at 10 m isotropic spacing.
It lays a 5500 m/s basement and a 30 cell pre-carbonate at 4000 m/s with
sinusoidal interbeds.
The carbonate is added via two \carbonateplatform passes: a 120 cell main platform with three
lateral platform centers (core 5500, forereef 5000, lagoon 4200, carbonate
basin 3500 m/s) and a 40 cell patch-reef pass with four smaller reef
bodies.
A 36 cell basin-fill deposit at 3300 m/s drapes the carbonate flanks.
Seven \deposit events then reproduce the clastic overburden in stratigraphic
order from Lower Mississauga (3800 m/s) up through Banquereau Upper (1900
m/s), with a thin Wyandot chalk unit (3400 m/s, suppressed interbed taper)
embedded in the middle.
A final 20 cell shallow Pleistocene deposit at 1700 m/s drapes the faulted
surface before the water column.
Three \squish events apply basin-scale vertical warping at decreasing amplitude (90, 65, 40 m).
One listric main fault (azimuth $50^\circ$, dip $30^\circ$, slip fraction
0.025) and four V-shaped secondaries (paired at $25^\circ$ and $155^\circ$
around common centers) give a total of five fault events.
A two-body salt diapir is placed on the right edge (placement bounds $x \in [0.84, 0.98]$),
vertically elongated (correlation anisotropy $[1.5, 0.6, 0.6]$), and
flanked by the \saltwedge module.
A 13 cell water column at 1480 m/s caps the model.

What is real: the carbonate-clastic contrast, the
five faults, the small-edge diapir, and the broad basin-warping sequence.
What we skip: the specific Abenaki reef-margin mode
(rimmed-shelf versus barrier-bank) and the Penobscot Canyon erosional feature.

\subsection{F3}
\label{app:model_details:f3}

Our F3 model is a deltaic-shelf model with deep salt and large-scale
faulting.
The reference is the F3 Block in the Dutch sector of the Central
Graben, an open 3D seismic volume widely used in seismic interpretation
\citep{silva2019netherlands}.
Two features drive the modeling choices.
First, Permian-age salt (the Zechstein) sits at depth and acts as a weak layer:
sediment loading above it generates large-scale faults that sole into
the salt rather than into the underlying brittle basement
\citep{davison2000central}.
Second, the Plio-Pleistocene section records the
Eridanos river system, a continental-scale delta that drained northwestern
Europe into the southern North Sea and left behind repeated basinward-
stepping clinothem packages (a clinothem is one dipping shelf-slope-basin
sediment body) \citep{overeem2001eridanos, kuhlmann2006integrated}.

Build-time grid is $1000 \times 1000 \times 1200$ at 20 m horizontal and
2.5 m vertical spacing; the shipped volumes resample to a $619 \times 1000 \times 1000$ grid at 10 m resolution. The stratigraphy is built deep
to shallow: a 3450 m/s basement, a pre-chalk Scruff analogue (2600 m/s), a
Chalk unit with the lowest interbed contrast in the model (2900 m/s, 40 m/s
std), a Lower Tertiary package (2200 m/s), and a chaotic mass-transport
layer (2300 m/s, 250 m/s std: the \emph{highest} interbed contrast in the
model). A rifting-phase squish (75 m max shift at $90^\circ$ azimuth) warps
the pre-clinoform section. The delta itself is built by the
\deltaclinoformdeposit module: ten forestepping clinothems (range
8--12 per realization), twelve beds each on average, stacked along a
$133^\circ$ SE trajectory with a rollover at 18\,\% along the profile, a
flat topset fraction of 0.5, and a toe-flattening strength of 0.75. A
second broad squish (35 m max shift, near-zero azimuth) warps the
pre-salt section before upper deposits are applied. Four shallower
deposits (Purple 1925, Upper Deep 1850, Upper Mid 1800, Upper Shallow
1750 m/s) overlie the delta. One main long-wavelength fault
(azimuth $145^\circ$, dip $30^\circ$), a left-side fault cluster of
$\sim$14 faults, and 2--4 geologically-placed support faults (offset from
the main fault by $+15^\circ$ and $+165^\circ$, 60/40 split) reproduce the
F3 fault population. A Yellow-Pleistocene cap at 1650 m/s drapes the
faulted surface, and a 30 cell water column at 1500 m/s tops the model.
Five salt-body groups (three on a diagonal chain, two outskirt) are
placed below $z = 0.40 \cdot n_z$ at 3500 m/s, then draped by a
surface-conforming \saltwedge pass.

We refer to the shelf-slope-basin geometry as
\emph{oblique-tangential} using the seismic-stratigraphy vocabulary of
\citet{mitchum1977stratigraphic}, and describe the stacking pattern as
\emph{forestepping} or via the shelf-edge-trajectory language of
\citet{hellandhansen2009trajectory}.
The word ``progradation'' describes a
temporal depositional process; our model is a static velocity volume, so
the geometric terms fit better.

What is real: the clinothem geometry, the Zechstein-decoupled fault style,
and the deep-to-shallow velocity trend. What we skip: intra-clinothem
mass-transport deposits, salt-pillar welding into the Jurassic section, and
glaciomarine fabrics in the uppermost Pleistocene.

\subsection{Fault}
\label{app:model_details:fault}

The Fault setting is a stress test, not a field.
It gives the training set dense normal-fault populations with high velocity contrast
across sub-parallel fault arrays.
The closest natural analogue is the post-rift section of the North Sea (Viking and Central Graben), where two superposed Mesozoic rifting phases left a multi-orientation normal-fault
population with widespread block rotation
\citep{badley1988structural, underhill1993jurassic}.
We reference the fault-population literature for the mechanism by which primary faults
seed sub-seismic secondaries: displacement--length scaling
\citep{cowie1992displength}, 3D fault-surface geometry
\citep{nicol1996shapes}, relay-ramp breaching between overstepping
segments \citep{peacock1994relay}, and along-strike displacement patterns
from the East African Rift \citep{morley1999patterns}.
The integrating tectono-sedimentary framework is described in \citet{gawthorpe2000tectono}.

\begin{figure}[t]
    \centering
    \includegraphics[width=\linewidth]{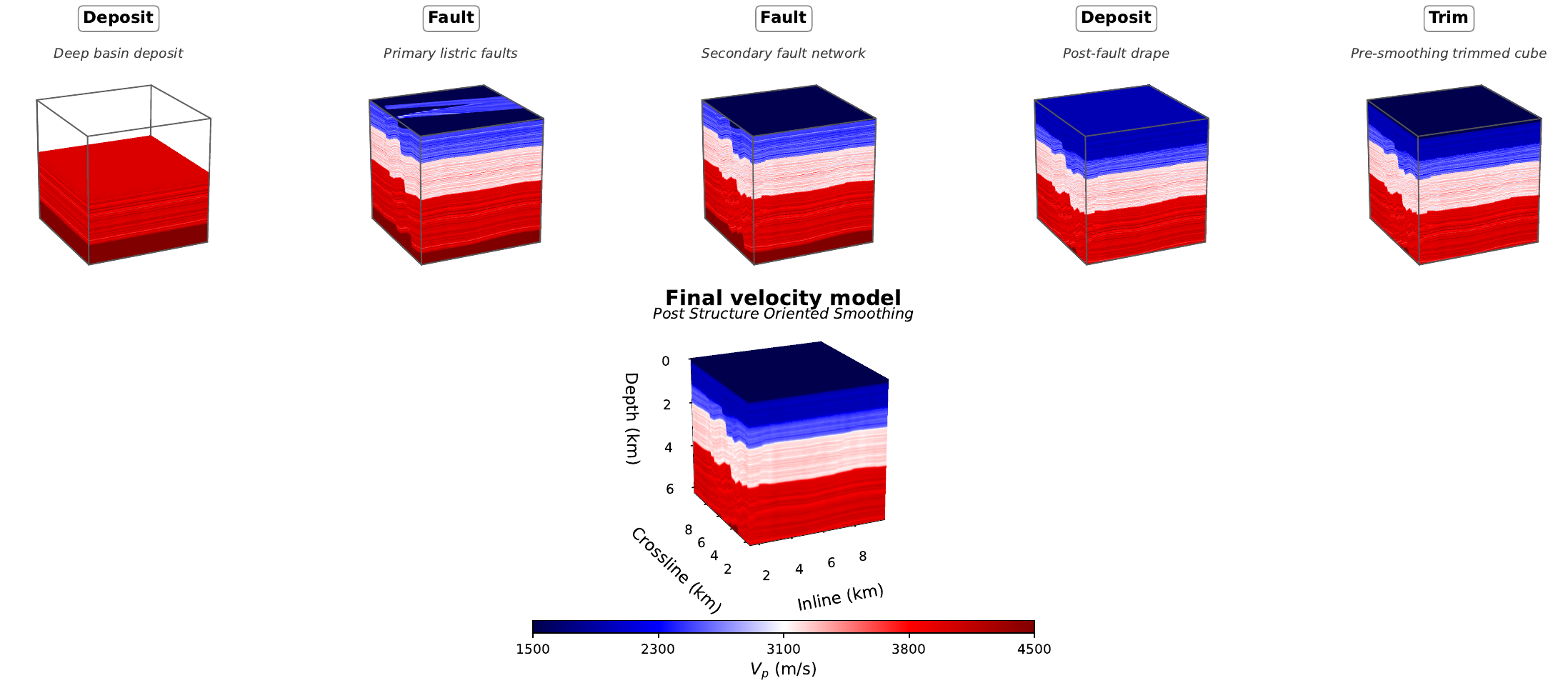}
    \caption{Per-stage build of one realization of the Fault
        setting.
        Reading left-to-right, top-to-bottom: the basement (5000 m/s) is
        laid down, followed by the four contrast-producing deposits
        (4000, 3200, 2500 m/s and the post-fault drape at 1950 m/s)
        interleaved with two squish events that warp the lower stack.
        The drape deposit then heals the discontinuity-bearing surface
        and the water column caps the model.}
    \label{fig:fault_flow}
\end{figure}

The builder stacks four contrast-producing deposits on a 5000 m/s basement
(\cref{fig:fault_flow}): deep 4000 m/s, mid 3200 m/s, upper 2500 m/s, and a post-fault drape at
1950 m/s. Two squish events interrupt the stack (200 m and 130 m max
shift). Each realization then draws \emph{3, 4, 5, or 6} primary faults
from an inclusive integer range; primary count is \emph{not} fixed. All
primaries share one system azimuth drawn from $[-45^\circ, +45^\circ]$,
are confined to non-crossing lanes perpendicular to that azimuth, and
carry the same slip direction so the array reads as a single sub-parallel
population. Per primary, 20--30 secondary faults are generated using
Riedel-shear geometry: synthetic R-shears offset by $+15^\circ$ from the
primary, antithetic R$'$-shears by $+165^\circ$, with 85\,\% synthetic
bias and a 0.75 hanging-wall placement bias. A minimum-distance filter
rejects overlapping subsidiaries, so the post-filter secondary count
varies with realization. After faulting, a drape deposit heals the
discontinuity-bearing surface, a 15 cell water column at 1480 m/s goes on
top, and the final clamp to $[1500, 5000]$~m/s removes any outliers
introduced by fault-zone interpolation.

What is real: the primary-array style, the Riedel-geometry secondary
pattern, and the high-contrast deposit stack.
What we skip:
overpressure-driven polygonal faulting, fault-zone damage halos, and
listric detachment into a ductile basal layer.
These realizations push the wavefield operator against dense
discontinuities, not against the mechanical specifics of any named rift
system.

\subsection{Gulf of Mexico}
\label{app:model_details:gom}

Our GoM analogue is a deepwater salt-tectonics model: multiple salt bodies
embedded in a layered sediment column, with sediments deformed upward
against salt flanks.
The setting is the northern Gulf of Mexico passive
margin, where salt from a deep Jurassic source layer has risen through the
overlying sediment as walls, stocks, sheets, and coalesced canopies
\citep{diegel1995cenozoic, peel1995genetic, hudec2013jurassic}.
\emph{Allochthonous} means that the salt has been transported out of its
original stratigraphic position, leaving behind sediment-filled
\emph{minibasins} between the new salt bodies.
These minibasins form because sediment loading differentially depresses the surrounding salt \citep{hudec2007terra}.
The rise of a salt diapir also deforms the flanking sediments, producing thinned and folded
packages whose geometry is formalized as \emph{halokinetic sequences}
(narrow hook sequences when the diapir rises fast relative to
sedimentation, broad wedge sequences when it rises slowly)
\citep{giles2012halokinetic}.
The geometric consequence for imaging is a
large sediment-to-salt velocity jump that produces shadow zones and
multipathing in ray-based tomography \citep{sava2004wemva1, sava2004wemva2, jones2014seismic, jackson2017salt}.

\begin{figure}[t]
    \centering
    \includegraphics[width=\linewidth]{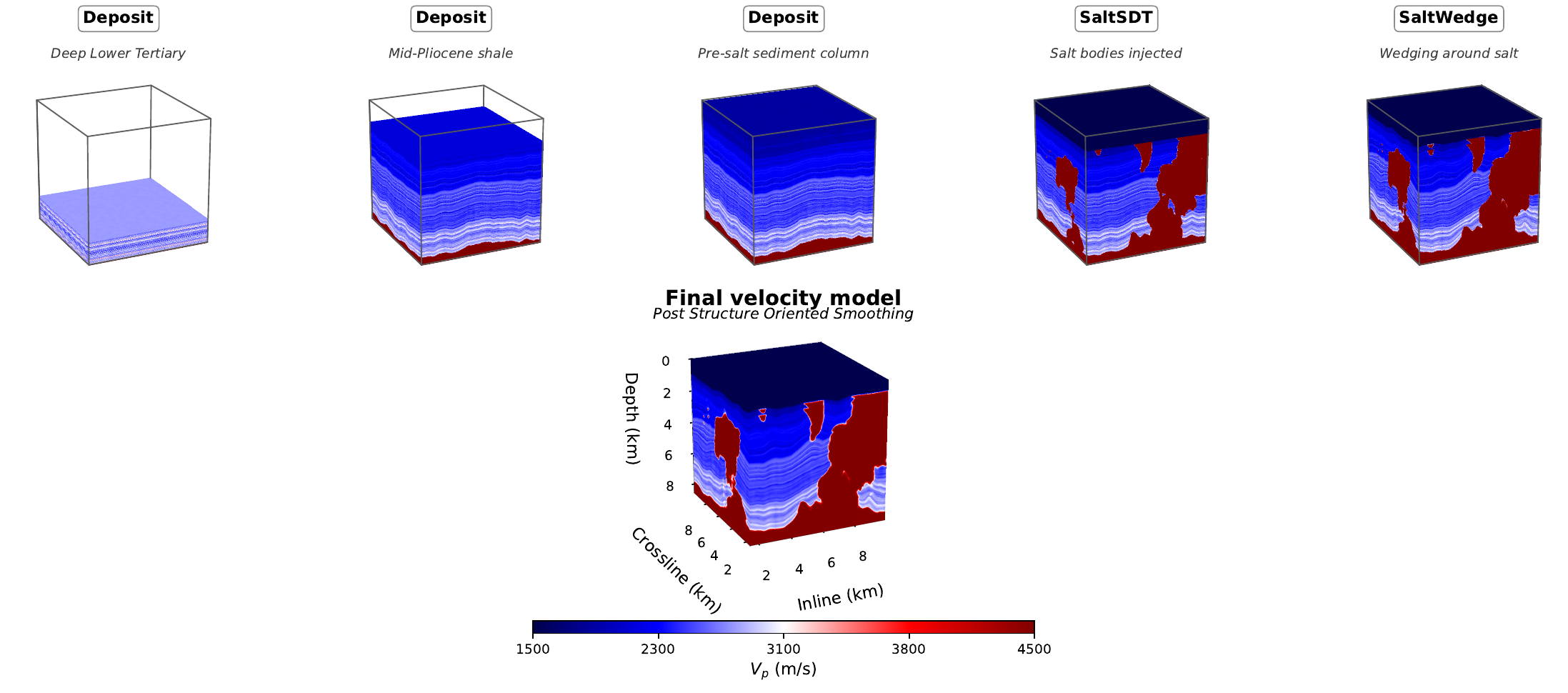}
    \caption{Per-stage build of one realization of the Gulf of Mexico setting.
        Reading left-to-right, top-to-bottom: the 5 cell basement
        (4500 m/s) is overlain by nine sediment deposits whose base
        velocities decrease upward (Lower Tertiary 2900 m/s through
        Pleistocene 1800 m/s), with five squish events of decreasing
        amplitude (500, 400, 300, 200, 100 m) interleaved between
        specific deposits to produce depth-dependent folding.
        The salt stage inserts SDT-generated bodies at depth fractions
        $[0.15, 0.60]$, and the wedging stage deforms the surrounding
        sediment to produce the upward onlap and drape characteristic
        of halokinetic sequences.}
    \label{fig:gom_flow}
\end{figure}

The builder runs on a $1000 \times 1000 \times 800$ grid at 10 m resolution.
A 5 cell 4500 m/s basement is overlain by nine \deposit events
(\cref{fig:gom_flow}) with base velocities decreasing upward from
2900 m/s (deep Lower Tertiary) to 1800 m/s (Pleistocene).
Five squish events interleave with the
deposits; squish amplitude decreases upsection from 500 m to 100 m,
giving the kind of depth-dependent folding one expects when older sediment
has seen more deformation.
The \saltsdt module inserts 2--7 salt
bodies per realization (base 4) at 4500 m/s, placed at depth fractions
$[0.15, 0.60]$, with a correlation length of 0.7 in each axis on the
GRF perturbing the body boundary.
The \saltwedge module then deforms the
surrounding sediment with a combination of a Gaussian weight in
distance to the nearest vertical flank, a radial drag weight raised
to the 1.8 power, a depth-dependent taper, and a fractal
spatial-amplitude field.
The net effect curves sediment upward against salt flanks and drapes it onto salt roofs, which is the geometric shape that the tapered (wedge) class of halokinetic sequence produces
\citep{giles2012halokinetic}.
A 96 cell water column at 1500 m/s caps the column.

What is real: the $\sim$2700 m/s salt-to-sediment velocity jump, the variety of
individual salt-body shapes that a parametric perturbation can span, and
the upward-flanking sediment geometry.
What we skip: welded or rafted
salt architectures, dissolution textures, and multi-stage salt-tectonic
histories.

\subsection{Salt Canopy}
\label{app:model_details:canopy}

The Salt Canopy model represents a second GoM setting.
Multiple allochthonous salt sheets have coalesced laterally into a shallow,
approximately tabular body above interbedded sediments.
Unlike the diapir-dominated GoM setting above, the canopy is continuous in map view
and produces the canonical subsalt imaging problem
\citep{hudec2006advance, pilcher2011primary, jones2014seismic,
    sava2004wemva2}.
The BP-operated Tiber discovery in Keathley Canyon block 102 is a widely cited field example of canopy salt.

The four canopy models in the dataset predate the \saltsdt
module and were built with a legacy code path (\texttt{pysaltmodel})
on a $1584 \times 2096 \times 536$ generation grid at 10 m resolution, then
resampled and depth-truncated to the common $619 \times 1000
    \times 1000$ grid.
Each salt body is the volume between two 2D fractal-noise elevation surfaces (a ``top'' and a
``bot'') filled at salt velocity, with a frequency-octave ladder of
$[2, 4, 8, 16, 32]$ and an amplitude ladder of $[1, 0.3, 0.1, 0.05, 0.025]$. Sediment is filled both below
and above the salt regime. A correction pipeline then inpaints low-velocity voxels,
caps velocities at 5000 m/s,
cuts below a detected basement, adds a $130 \pm 10$-cell water column at
1500 m/s, and clamps the final volume to $[1500, 5000]$~m/s.
This model building process is inspired by
\citet{farris2023learning,farris2023thesis}.

What is real: the tabular canopy morphology and the subsalt
sediment--salt--sediment contrast.
What we skip: inter-canopy saline-entrainment textures and sub-canopy overpressure signatures mapped
in the real-world Keathley Canyon section.

\subsection{SEAM Phase I}
\label{app:model_details:seam}

The twelve SEAM models are \emph{reused, not generated}.
They come from the SEG Advanced Modeling (SEAM) Phase I
3D earth model \citep{fehler2011seam}, which was built by an industry
consortium between 2007 and 2013 to model a salt canopy region of
the deepwater Gulf of Mexico.
We incorporate SEAM because it is
a community-validated reference model in the same regime as our Salt Canopy setting.
The Phase I model is distributed under CC BY 4.0 through the SEG SEAM
Open Data program \citep{seg2024seamopendata}.

The raw Phase I model is $1501 \times 4001 \times 3501$ at 10 m resolution
($15\text{~km} \times 40\text{~km} \times 35\text{~km}$).
Our pipeline extracts twelve
$1501 \times 1000 \times 1000$ subregions on a $4 \times 3$ grid.
We apply structure-oriented smoothing to each subregion ($\sigma_{\text{tensor}} = 8$ cells, smoothing-scale $= 8$ cells,
$\alpha = 0.3$, 10 conjugate gradient iterations) and clamp to $[1500, 5000]$~m/s.
Each subregion is depth-truncated to 619 cells, which is followed by structure-oriented smoothing with $\sigma_{\text{tensor}} = 8$, smoothing-scale $= 8$, $\alpha = 0.3$, and 10 conjugate gradient iterations.

What is real: SEAM is based on real-world geology. What we skip: N/A.
\section{Velocity Model Builder: GPU vs. CPU Timing}
\label{app:builder_benchmark}

This appendix backs \cref{tab:builder_speedup} with hardware, methodology, per-stage timings, and per-module timing totals.
The modules are described in \cref{app:model_builder_modules}; the per-setting build sequences are in \cref{app:model_details}.

\paragraph{Hardware.}
A single compute node with one 80 GB A100 NVIDIA GPU and an AMD EPYC 7763 host (16-core / 256 GB allocation).
The GPU was held in CUDA Exclusive\_Process mode so no other process contended for the device.
CPU runs set \texttt{CUDA\_VISIBLE\_DEVICES=""} to avoid creating a CUDA context, with BLAS / OpenMP thread counts pinned to 16.

\paragraph{Methodology.}
Each (setting, device) pair runs as a new Python subprocess.
Per-stage timings use \texttt{time.perf\_counter()} bracketed by \texttt{torch.cuda.synchronize()} on the GPU run.
GPU utilization and memory usage are sampled 10 times per second, i.e., at 10 Hz.

\paragraph{Per-setting totals.}
\cref{tab:builder_total_breakdown} stacks the wall-time breakdown alongside peak GPU memory and peak CPU memory.
Wall is the full subprocess time including Taichi initialization, CUDA-context setup, and inter-stage cleanup; the gap between Wall and Build is the warm-up plus Python overhead.

\begin{table}[h]
    \centering
    \footnotesize
    \setlength{\tabcolsep}{5pt}
    \caption{Per-setting totals. Build is the sum of per-stage build times. Wall is the full subprocess wall time. Peak GPU (CPU)memory is the maximum GPU (CPU) memory used during the run.}
    \label{tab:builder_total_breakdown}
    \begin{tabular}{lrrrrrrr}
        \toprule
        Setting   & Stages & GPU build & GPU wall & Peak GPU    & CPU build & CPU wall & Peak CPU    \\
                  &        & (min)     & (min)    & memory (GB) & (min)     & (min)    & memory (GB) \\
        \midrule
        Penobscot & 21     & 5.57      & 6.66     & 48.6        & 92.67     & 92.72    & 74.2        \\
        F3        & 15     & 11.49     & 12.60    & 60.0        & 82.29     & 82.46    & 137.4       \\
        Fault     & 12     & 5.84      & 6.56     & 67.7        & 156.57    & 156.69   & 68.0        \\
        GoM       & 19     & 6.70      & 7.84     & 52.3        & 100.21    & 100.31   & 92.8        \\
        \bottomrule
    \end{tabular}
\end{table}

\paragraph{Per-stage breakdowns.}
\cref{tab:builder_per_stage_penobscot,tab:builder_per_stage_f3,tab:builder_per_stage_ifc,tab:builder_per_stage_gom} list every module call. \texttt{canonical\_sos} is the final \structuralsmoother call; the four \texttt{water} stages set the water column velocity and run in milliseconds on both CPU and GPU.

\begin{table}[h]
    \centering
    \scriptsize
    \caption{Per-stage build for the Penobscot setting. Speedup is the ratio of CPU to GPU build time. Avg / pk util are weighted-average and peak GPU utilization sampled at 10\,Hz across the stage. Pk CPU memory is the maximum sampled CPU memory usage.}
    \label{tab:builder_per_stage_penobscot}
    \begin{tabular}{rlrrrrrrr}
        \toprule
        \#                        & Stage                              & GPU s   & CPU s   & Speedup & Pk GPU      & Avg util & Pk util & Pk CPU      \\
                                  &                                    & build   & build   &         & memory (GB) & \%       & \%      & memory (GB) \\
        \midrule
        1                         & \texttt{basement}                  & 0.003   & 0.033   & 10.77x  & 0.08        & 0.0      & 0       & 0.58        \\
        2                         & \texttt{pre\_carbonate}            & 0.822   & 2.924   & 3.55x   & 8.55        & 2.2      & 14      & 8.98        \\
        3                         & \texttt{abenaki\_carbonate}        & 2.765   & 85.426  & 30.90x  & 10.32       & 38.1     & 100     & 11.01       \\
        4                         & \texttt{basin\_fill}               & 26.330  & 500.397 & 19.00x  & 21.53       & 61.2     & 69      & 22.20       \\
        5                         & \texttt{lower\_mississauga}        & 20.395  & 389.918 & 19.12x  & 21.34       & 65.1     & 74      & 22.03       \\
        6                         & \texttt{squish\_basin\_subsidence} & 0.732   & 14.626  & 19.98x  & 11.48       & 6.6      & 53      & 9.41        \\
        7                         & \texttt{upper\_mississauga}        & 22.446  & 443.507 & 19.76x  & 23.92       & 64.9     & 100     & 24.63       \\
        8                         & \texttt{logan\_canyon}             & 33.688  & 684.111 & 20.31x  & 25.65       & 66.8     & 73      & 26.35       \\
        9                         & \texttt{squish\_mid\_cret}         & 0.616   & 19.470  & 31.63x  & 17.62       & 24.6     & 43      & 14.55       \\
        10                        & \texttt{dawson\_canyon}            & 16.508  & 330.979 & 20.05x  & 27.04       & 66.9     & 75      & 27.76       \\
        11                        & \texttt{wyandot\_chalk}            & 4.055   & 84.690  & 20.88x  & 15.57       & 60.0     & 73      & 16.12       \\
        12                        & \texttt{banquereau\_lower}         & 18.614  & 381.580 & 20.50x  & 26.97       & 67.8     & 76      & 27.65       \\
        13                        & \texttt{banquereau\_upper}         & 17.934  & 347.060 & 19.35x  & 28.13       & 65.6     & 82      & 28.85       \\
        14                        & \texttt{squish\_cenozoic}          & 1.012   & 26.957  & 26.64x  & 25.13       & 29.0     & 44      & 20.55       \\
        15                        & \texttt{faults}                    & 26.256  & 990.567 & 37.73x  & 48.56       & 90.8     & 100     & 39.29       \\
        16                        & \texttt{final\_deposit}            & 7.229   & 150.028 & 20.76x  & 21.52       & 64.5     & 75      & 22.00       \\
        17                        & \texttt{water}                     & 0.000   & 0.004   & 7.95x   & 5.49        & 0.0      & 0       & 0.00        \\
        18                        & \texttt{salt}                      & 34.370  & 33.035  & 0.96x   & 14.71       & 0.6      & 15      & 15.91       \\
        19                        & \texttt{wedging}                   & 76.354  & 75.358  & 0.99x   & 30.84       & 2.4      & 100     & 74.18       \\
        20                        & \texttt{trim}                      & 0.159   & 1.283   & 8.08x   & 17.82       & 0.0      & 0       & 18.67       \\
        21                        & \texttt{canonical\_sos}            & 24.097  & 998.233 & 41.42x  & 40.35       & 95.0     & 100     & 49.73       \\
        \midrule
        \multicolumn{2}{l}{Total} & 334.38                             & 5560.18 & 16.63x  & 48.56   & 47.7        & 100      & 74.18                 \\
        \bottomrule
    \end{tabular}
\end{table}

\begin{table}[h]
    \centering
    \scriptsize
    \caption{Per-stage build for the F3 setting. Speedup is the ratio of CPU to GPU build time. Avg / pk util are weighted-average and peak GPU utilization sampled at 10\,Hz across the stage. Pk CPU memory is the maximum sampled CPU memory usage.}
    \label{tab:builder_per_stage_f3}
    \begin{tabular}{rlrrrrrrr}
        \toprule
        \#                        & Stage                    & GPU s   & CPU s    & Speedup & Pk GPU      & Avg util & Pk util & Pk CPU      \\
                                  &                          & build   & build    &         & memory (GB) & \%       & \%      & memory (GB) \\
        \midrule
        1                         & \texttt{basement}        & 0.170   & 0.017    & 0.10x   & 0.03        & 0.0      & 0       & 0.00        \\
        2                         & \texttt{pre\_chalk}      & 47.365  & 270.485  & 5.71x   & 19.95       & 65.9     & 100     & 20.22       \\
        3                         & \texttt{chalk}           & 18.233  & 98.506   & 5.40x   & 21.57       & 59.9     & 90      & 21.73       \\
        4                         & \texttt{lower\_tertiary} & 36.707  & 210.725  & 5.74x   & 24.27       & 69.2     & 74      & 24.54       \\
        5                         & \texttt{chaotic}         & 28.003  & 146.240  & 5.22x   & 25.81       & 66.8     & 74      & 25.62       \\
        6                         & \texttt{squish\_rifting} & 0.846   & 11.690   & 13.82x  & 18.61       & 7.5      & 12      & 15.03       \\
        7                         & \texttt{clinoforms}      & 28.218  & 360.272  & 12.77x  & 14.81       & 73.3     & 100     & 15.65       \\
        8                         & \texttt{squish}          & 0.960   & 15.206   & 15.84x  & 23.76       & 47.6     & 100     & 19.43       \\
        9                         & \texttt{upper\_deposits} & 78.208  & 427.135  & 5.46x   & 36.36       & 70.5     & 100     & 36.62       \\
        10                        & \texttt{faults}          & 73.570  & 1907.790 & 25.93x  & 57.10       & 89.4     & 100     & 53.03       \\
        11                        & \texttt{final\_deposit}  & 25.849  & 143.507  & 5.55x   & 36.92       & 69.4     & 81      & 36.89       \\
        12                        & \texttt{water}           & 0.001   & 0.006    & 7.41x   & 8.64        & 0.0      & 0       & 0.00        \\
        13                        & \texttt{salt}            & 239.939 & 190.560  & 0.79x   & 27.43       & 0.9      & 78      & 33.18       \\
        14                        & \texttt{wedging}         & 73.038  & 81.806   & 1.12x   & 44.53       & 2.9      & 100     & 137.43      \\
        15                        & \texttt{canonical\_sos}  & 38.288  & 1073.629 & 28.04x  & 60.00       & 95.0     & 100     & 108.32      \\
        \midrule
        \multicolumn{2}{l}{Total} & 689.40                   & 4937.57 & 7.16x    & 60.00   & 41.6        & 100      & 137.43                \\
        \bottomrule
    \end{tabular}
\end{table}

\begin{table}[h]
    \centering
    \scriptsize
    \caption{Per-stage build for the Fault setting. Speedup is the ratio of CPU to GPU build time. Avg / pk util are weighted-average and peak GPU utilization sampled at 10\,Hz across the stage. Pk CPU memory is the maximum sampled CPU memory usage.}
    \label{tab:builder_per_stage_ifc}
    \begin{tabular}{rlrrrrrrr}
        \toprule
        \#                        & Stage                       & GPU s   & CPU s    & Speedup & Pk GPU      & Avg util & Pk util & Pk CPU      \\
                                  &                             & build   & build    &         & memory (GB) & \%       & \%      & memory (GB) \\
        \midrule
        1                         & \texttt{basement}           & 0.004   & 0.091    & 22.48x  & 0.80        & 5.0      & 5       & 0.00        \\
        2                         & \texttt{deep\_deposit}      & 68.242  & 773.786  & 11.34x  & 23.61       & 66.0     & 71      & 23.85       \\
        3                         & \texttt{squish\_deep}       & 1.011   & 12.458   & 12.32x  & 14.03       & 1.8      & 3       & 11.65       \\
        4                         & \texttt{mid\_deposit}       & 49.081  & 404.522  & 8.24x   & 27.47       & 71.1     & 100     & 27.88       \\
        5                         & \texttt{squish\_shallow}    & 0.977   & 17.346   & 17.75x  & 20.82       & 0.0      & 0       & 16.82       \\
        6                         & \texttt{upper\_deposit}     & 30.110  & 243.264  & 8.08x   & 30.08       & 69.4     & 99      & 30.75       \\
        7                         & \texttt{primary\_faults}    & 49.585  & 1837.841 & 37.06x  & 67.67       & 97.1     & 100     & 53.36       \\
        8                         & \texttt{secondary\_faults}  & 91.252  & 4636.105 & 50.81x  & 55.20       & 72.1     & 100     & 56.01       \\
        9                         & \texttt{post\_fault\_drape} & 37.125  & 730.076  & 19.67x  & 37.80       & 67.6     & 75      & 38.40       \\
        10                        & \texttt{water}              & 0.001   & 0.010    & 10.09x  & 8.99        & 0.0      & 0       & 0.00        \\
        11                        & \texttt{trim\_clamp}        & 0.128   & 2.369    & 18.53x  & 21.70       & 0.0      & 0       & 22.12       \\
        12                        & \texttt{canonical\_sos}     & 23.130  & 736.401  & 31.84x  & 41.55       & 95.7     & 100     & 68.04       \\
        \midrule
        \multicolumn{2}{l}{Total} & 350.64                      & 9394.27 & 26.79x   & 67.67   & 74.7        & 100      & 68.04                 \\
        \bottomrule
    \end{tabular}
\end{table}

\begin{table}[h]
    \centering
    \scriptsize
    \caption{Per-stage build for the Gulf of Mexico setting. Speedup is the ratio of CPU to GPU build time. Avg / pk util are weighted-average and peak GPU utilization sampled at 10\,Hz across the stage. Pk CPU memory is the maximum sampled CPU memory usage.}
    \label{tab:builder_per_stage_gom}
    \begin{tabular}{rlrrrrrrr}
        \toprule
        \#                        & Stage                                & GPU s   & CPU s    & Speedup & Pk GPU      & Avg util & Pk util & Pk CPU      \\
                                  &                                      & build   & build    &         & memory (GB) & \%       & \%      & memory (GB) \\
        \midrule
        1                         & \texttt{basement}                    & 0.254   & 0.024    & 0.09x   & 0.04        & 0.0      & 0       & 0.00        \\
        2                         & \texttt{deep\_lower\_tertiary}       & 48.448  & 898.372  & 18.54x  & 19.89       & 60.4     & 69      & 20.56       \\
        3                         & \texttt{squish\_1}                   & 0.461   & 10.536   & 22.85x  & 7.51        & 5.0      & 5       & 6.64        \\
        4                         & \texttt{mid\_tertiary}               & 34.037  & 702.630  & 20.64x  & 23.25       & 66.7     & 73      & 24.09       \\
        5                         & \texttt{upper\_miocene\_pliocene}    & 38.886  & 775.983  & 19.96x  & 25.27       & 65.5     & 75      & 25.84       \\
        6                         & \texttt{squish\_2}                   & 0.656   & 19.518   & 29.74x  & 18.45       & 0.0      & 0       & 15.15       \\
        7                         & \texttt{lower\_pliocene\_basal}      & 32.194  & 642.655  & 19.96x  & 28.12       & 68.3     & 81      & 28.93       \\
        8                         & \texttt{lower\_pliocene\_upper}      & 13.520  & 279.740  & 20.69x  & 28.65       & 66.3     & 72      & 29.18       \\
        9                         & \texttt{squish\_3}                   & 1.081   & 26.369   & 24.39x  & 24.49       & 9.0      & 33      & 20.02       \\
        10                        & \texttt{mid\_pliocene\_shale}        & 17.827  & 369.199  & 20.71x  & 30.80       & 68.8     & 100     & 31.69       \\
        11                        & \texttt{upper\_pliocene\_transition} & 7.931   & 169.656  & 21.39x  & 24.37       & 64.9     & 72      & 25.25       \\
        12                        & \texttt{squish\_4}                   & 0.834   & 29.767   & 35.70x  & 29.02       & 28.8     & 64      & 23.42       \\
        13                        & \texttt{upper\_pliocene\_reservoir}  & 22.693  & 430.846  & 18.99x  & 33.10       & 67.4     & 100     & 33.96       \\
        14                        & \texttt{pleistocene}                 & 8.617   & 181.304  & 21.04x  & 24.74       & 65.8     & 72      & 25.59       \\
        15                        & \texttt{squish\_5}                   & 0.900   & 31.409   & 34.89x  & 32.39       & 28.9     & 100     & 26.31       \\
        16                        & \texttt{water}                       & 0.000   & 0.016    & 45.31x  & 6.87        & 0.0      & 0       & 0.00        \\
        17                        & \texttt{salt\_bodies}                & 45.527  & 44.350   & 0.97x   & 30.32       & 0.7      & 68      & 19.92       \\
        18                        & \texttt{wedging}                     & 94.578  & 96.718   & 1.02x   & 38.62       & 2.8      & 100     & 92.80       \\
        19                        & \texttt{canonical\_sos}              & 33.502  & 1303.356 & 38.90x  & 52.31       & 97.3     & 100     & 58.67       \\
        \midrule
        \multicolumn{2}{l}{Total} & 401.95                               & 6012.45 & 14.96x   & 52.31   & 45.5        & 100      & 92.80                 \\
        \bottomrule
    \end{tabular}
\end{table}

\paragraph{Per-module totals and observations.}
\cref{tab:builder_class_aggregate} groups the per-stage rows by builder module.
The largest speedups are on \fault (38.9$\times$) and \structuralsmoother (34.5$\times$).
\deposit has a smaller per-call speedup (13.6$\times$) but its high call count makes it the biggest contributor to total time saved.
\saltsdt and \saltwedge show no GPU advantage; improving these implementations is a target for future work.

\begin{table}[h]
    \centering
    \small
    \caption{Per-module totals over all four settings. \textit{n} is the number of times each stage is invoked; \textit{GPU s} and \textit{CPU s} are the summed times for each stage. \textit{Avg util} is the average GPU utilization (sampled at 10 Hz). \textit{Peak GPU memory} is the maximum GPU memory used by the module.}
    \label{tab:builder_class_aggregate}
    \begin{tabular}{lrrrrrr}
        \toprule
        Module                 & \textit{n} & GPU s  & CPU s    & Speedup       & Avg util \% & \shortstack{Peak GPU \\ memory (GB)} \\
        \midrule
        \deposit               & 27         & 773.15 & 10480.82 & 13.56$\times$ & 66.7        & 36.92                \\
        \saltsdt               & 3          & 319.84 & 267.94   & 0.84$\times$  & 0.9         & 30.32                \\
        \saltwedge             & 3          & 243.97 & 253.88   & 1.04$\times$  & 2.7         & 44.53                \\
        \fault                 & 4          & 240.66 & 9372.30  & 38.94$\times$ & 84.6        & 67.67                \\
        \structuralsmoother    & 4          & 119.02 & 4111.62  & 34.55$\times$ & 95.8        & 60.00                \\
        \deposit (drape)       & 1          & 37.12  & 730.08   & 19.67$\times$ & 67.6        & 37.80                \\
        \deltaclinoformdeposit & 1          & 28.22  & 360.27   & 12.77$\times$ & 73.3        & 14.81                \\
        \squish                & 12         & 10.09  & 235.35   & 23.33$\times$ & 16.4        & 32.39                \\
        \carbonateplatform     & 2          & 3.59   & 88.35    & 24.63$\times$ & 29.9        & 10.32                \\
        Basement seed          & 4          & 0.43   & 0.16     & 0.38$\times$  & 0.0         & 0.80                 \\
        Trim/clamp             & 2          & 0.29   & 3.65     & 12.74$\times$ & 0.0         & 21.70                \\
        Water column           & 4          & 0.00   & 0.04     & 13.45$\times$ & 0.0         & 8.99                 \\
        \bottomrule
    \end{tabular}
\end{table}

\section{Seismic Forward Modeling Details}
\label{app:seismic_forward}

This appendix gives slice extraction, numerical scheme, source wavelet, and acquisition geometry details for the seismic data generation in \cref{sec:seismic_generation}.

\subsection{Slice extraction and dataset organization}
\label{app:seismic_forward:slices}

The 2D dataset is produced from the 42 3D volumes (\cref{tab:model_inventory})
by a deterministic extractor with seed 42. All volumes are first
depth-truncated to 619 cells (the minimum depth across the 10 GoM volumes),
then re-smoothed via the \structuralsmoother module at the truncated
resolution so that the smoothing scale is consistent across settings.
Slice planes are placed with a 10\,\% boundary margin on each horizontal
axis (to avoid boundary-layer artifacts) and an orientation split of 50\,\%
inline vs 50\,\% crossline per setting, sampled without replacement at
integer slice indices. Inline and crossline indices are drawn from
independent random streams (per-model seeds 42 + model\_index and 42 +
model\_index + 10000 respectively), and the SEAM volumes are extracted as
12 sub-regions of $1000 \times 1000$ cells with 79 slices per sub-region.
Before slicing, each of the 42 3D velocity volumes is normalized in place
to fit $v_p \in [1500, 5000]$~m/s: a linear rescale from $[v_{\min},
            v_{\max}]$ to $[v_{\min}, 5000]$ when $v_{\max} > 5000$ (preserving every
relative velocity contrast within the volume), followed by a hard clip
floor at $1500$. All extracted 2D slices inherit this normalization, so
the 4{,}276 shipped slices satisfy $v_p \in [1500, 5000]$~m/s without
further per-slice modification, consistent with the dispersion
(\cref{sec:dispersion}) and CFL (\cref{sec:cfl}) constraints derived below.

\paragraph{Splits.}
Per-setting train slice counts: F3 $10 \times 97 = 970$, GoM $10 \times 73 = 730$,
faulted-complex $5 \times 145 = 725$, salt-canopy $4 \times 181 = 724$,
SEAM $12 \times 79 = 947$. The training split is the union of these
slices, $970 + 730 + 725 + 724 + 947 = 4{,}096$. Twenty additional slices
per training setting form a 100-slice in-distribution test split drawn
from non-overlapping indices within the same volumes. The 80 Penobscot
slices form the out-of-distribution test split, giving a dataset-wide
total of $4{,}096 + 100 + 80 = 4{,}276$ 2D slices.

\paragraph{Parquet index.}
The benchmark root indexes 47{,}078 rows total: 42 3D models, 4{,}276
slices, 21{,}380 wavefields ($4{,}276 \times 5$ frequency bands), and
21{,}380 shot-gather cubes. Each row carries 25 columns, including
\texttt{slice\_id}, \texttt{model\_id}, \texttt{data\_type}
(\texttt{model}, \texttt{slice}, \texttt{wavefield}, or \texttt{gather}),
\texttt{model\_type}, \texttt{split}, \texttt{file\_path},
\texttt{orientation}, \texttt{frequency\_band}, source coordinates, and
precomputed per-slice velocity statistics. Slices and their paired
wavefields and shot-gather cubes share the same \texttt{slice\_id}, so
joining across \texttt{data\_type} on \texttt{slice\_id} materializes the
paired triplets at any given frequency band.

\subsection{Continuous PDE and boundary conditions}
\label{app:seismic_forward:pde}

The forward simulator solves the 2D acoustic, constant-density, isotropic
wave equation in the $(x, z)$ plane. With $p(x, z, t)$ the pressure
wavefield, $v_p(x, z)$ the velocity model, and $s(x, z, t)$ the source
term (a band-limited Ricker pulse, \cref{sec:wavelet}),
\begin{equation}
    \frac{1}{v_p(x, z)^2} \frac{\partial^2 p}{\partial t^2}
    - \left( \frac{\partial^2 p}{\partial x^2} + \frac{\partial^2 p}{\partial z^2} \right)
    = s(x, z, t).
    \label{eq:acoustic_2d}
\end{equation}
The factor $1/v_p^2$ scales the temporal acceleration term: lower-velocity
regions have a larger coefficient on $\partial_t^2 p$, so they admit
slower propagation. The Laplacian $\partial_x^2 + \partial_z^2$ measures
spatial curvature of the wavefield, which is the restoring force that
drives propagation.

\paragraph{Initial conditions.}
The pressure and its time derivative are zero everywhere at $t = 0$,
$p(x, z, 0) = 0$ and $\partial_t p(x, z, 0) = 0$. The wavefield is excited
entirely by the source term $s$, which fires a localized Ricker pulse
near the surface (\cref{sec:acquisition}).

\paragraph{Free surface at the top.}
The top edge of the simulation domain ($z = 0$) is a free surface,
modeling the air--water (or air--rock) interface where pressure must
vanish:
\begin{equation}
    p(x, 0, t) = 0, \qquad t \in [0, T].
    \label{eq:free_surface}
\end{equation}
This is enforced in Devito by the \texttt{fs=True} setting on the
\texttt{Model} object \citep{louboutin2019devito} and produces the
sign-flipped reflection of upgoing energy back into the model that is
characteristic of marine acquisitions.

\paragraph{Cerjan-style sponge damping on the other three sides.}
The left, right, and bottom edges are absorbing rather than physical:
the simulation grid is just a truncation of an unbounded earth model,
and we want waves arriving at those edges to dissipate without spurious
reflection. Devito's \texttt{bcs="damp"} configuration adds a 60-cell
sponge layer along each absorbing side, following the formulation of
\citet{cerjan1985nonreflecting}: inside the sponge, the wave equation
acquires a first-order temporal-damping term proportional to a
position-dependent damping coefficient $\eta(\mathbf{x})$,
\begin{equation}
    \frac{1}{v_p(\mathbf{x})^2} \frac{\partial^2 p}{\partial t^2}
    - \nabla^2 p
    + \eta(\mathbf{x})\, \frac{\partial p}{\partial t}
    = s(\mathbf{x}, t).
    \label{eq:cerjan_pde}
\end{equation}
The coefficient is zero throughout the interior, recovering
\eqref{eq:acoustic_2d} there, and rises smoothly to its peak at the
outer edge. With $n_\text{bl}$ damping cells, $h$ the grid spacing, and
$\rho \in [0, 1]$ the normalized distance into the sponge,
\begin{equation}
    \eta(\mathbf{x}) = \frac{c_\text{damp}}{h} \left[ \rho - \frac{\sin(2\pi \rho)}{2\pi} \right],
    \qquad c_\text{damp} = \frac{1.5\,\ln(1/R)}{n_\text{bl}},
    \label{eq:cerjan_damp}
\end{equation}
where $R = 10^{-3}$ is the target reflection coefficient at the outer
edge ($-60$\,dB). The $\sin(2\pi \rho)/(2\pi)$ term tapers the rate of
growth so $\eta$ rises gradually rather than as a hard step, which is
what keeps the damping itself from generating new spurious reflections.
With $n_\text{bl} = 60$ and $h = 10$\,m, the sponge thickness is
$0.6$\,km --- wide enough at our highest band ($\fmax = 25$\,Hz) for
waves to traverse the layer over multiple wavelengths and so be
attenuated to the design tolerance before reaching the outer boundary.

\subsection{Solving the Wave Equation Numerically}
\label{sec:pde_discretization}

\subsubsection{Spatial Sampling (Dispersion Control)}
\label{sec:dispersion}

Finite-difference schemes propagate short-wavelength components too slowly relative to the true phase velocity, an artifact known as numerical dispersion \citep{liu2009new}. To keep this error below a prescribed tolerance, the grid must resolve the shortest wavelength $\lmin = \vmin / \fmax$ with at least $G$ grid points:
\begin{equation*}
    h \leq \frac{\vmin}{G \fmax}.
\end{equation*}

The required $G$ depends on the stencil order: higher-order schemes need
fewer points per wavelength for the same accuracy. For our 8th-order
spatial stencil, $G = 4$ keeps the phase-velocity dispersion error well
below 1\,\% over the bandwidths we simulate, so the grid spacing is
fixed by the highest-frequency content propagating through the slowest
part of the model.

\subsubsection{CFL Stability Condition}
\label{sec:cfl}

Once $h$ is chosen, the time step cannot be chosen independently. The Courant--Friedrichs--Lewy (CFL) condition provides the maximum $\Delta t$ for which the explicit time-marching scheme remains stable \citep{courant1928uber}:
\begin{equation*}
    \Delta t \leq \smax \frac{h}{\vmax},
\end{equation*}
where $\smax = \left( \sqrt{d \sum_{m=1}^{M} |a_m|} \right)^{-1}$ is the
maximum stable Courant number, $d$ is the number of spatial dimensions,
and $\{a_m\}_{m = 1}^M$ are the finite-difference coefficients
\citep{fornberg1988generation}. Note the asymmetry: the dispersion
criterion involves $\vmin$ (shortest wavelength), while stability involves
$\vmax$ (fastest propagation). Both constrain us simultaneously.

\subsubsection{Temporal Subsampling (Nyquist)}
\label{sec:nyquist}

The CFL bound forces $\Delta t = 1$\,ms for stability, but the band-limited
wavefield does not need to be saved at that rate. Storing every
computational step would inflate the dataset by an order of magnitude
without adding any information that the band-limit cannot already
recover, and would slow training throughput in proportion. Shannon--Nyquist
sets the floor: a signal band-limited at $\fmax$ can be reconstructed
exactly from samples taken every $1/(2\fmax)$ seconds. We therefore save
every $k$-th computed step, where
\begin{equation*}
    k = \left\lfloor \frac{N}{2 \fmax T} \right\rfloor.
\end{equation*}
This yields $\Nsaved = \lceil N/k \rceil$ snapshots at an effective
sampling interval of $\dtsaved = k \cdot \Delta t$. We use $k = 14$
uniformly across all five frequency bands so the saved tensor has the
same temporal dimension regardless of source bandwidth; this is
conservative for the lower bands (\cref{tab:bands}) but yields a uniform
$(\Nsaved, n_z, n_x)$ shape that batches naturally during training.

\subsubsection{Configuration Summary}
\label{sec:config}

The central design constraint is that every simulation, regardless of
which frequency band the source wavelet targets, must produce a
wavefield tensor of identical shape. Neural operators expect fixed-size
inputs during training, and batching requires uniform dimensions. We
therefore adopt a single spatial grid, a single time step, and a single
temporal subsampling factor across all 21{,}380 wavefield simulations
and 21{,}380 shot-gather cube simulations (5 frequency bands $\times$
4{,}276 slices); \cref{tab:params} summarizes the resulting simulation parameters.

\begin{table}[ht]
    \centering
    \caption{Simulation parameters shared across all 5 frequency bands.
        Wavefields and shot-gather cubes share the same grid, time step,
        stencil order, boundary, and subsampling factor; they differ
        only in recording time, source geometry, and whether the full
        wavefield or a receiver-extracted gather is saved.}
    \label{tab:params}
    \smallskip
    \footnotesize
    \begin{tabular}{@{}llc@{}}
        \toprule
        Parameter                       & Symbol     & Value                                \\
        \midrule
        Grid spacing                    & $h$        & 10\,m (isotropic)                    \\
        Spatial FD order                & $2M$       & 8 ($M = 4$)                          \\
        Max P-wave velocity             & $\vmax$    & 5000\,m/s                            \\
        Min P-wave velocity             & $\vmin$    & 1500\,m/s                            \\
        Computational time step         & $\Delta t$ & 1.0\,ms                              \\
        Recording time (wavefield)      & $T$        & 5.0\,s                               \\
        Recording time (shot-gather)    & $T$        & 8.0\,s                               \\
        Total computational steps       & $N$        & 5001 (wavefield) / 8001 (gather)     \\
        Subsampling factor              & $k$        & 14                                   \\
        Saved snapshots (wavefield)     & $\Nsaved$  & 358                                  \\
        Saved time samples (gather)     & $\Nsaved$  & 572                                  \\
        Saved time interval             & $\dtsaved$ & 14.0\,ms                             \\
        Slice depth                     & $n_z$      & 619                                  \\
        Slice width                     & $n_x$      & 1000 (typical)                       \\
        Source wavelet                  & ---        & Bandpass Ricker (\cref{sec:wavelet}) \\
        Absorbing boundary              & ---        & 60-cell sponge damping; free top     \\
        \textbf{Wavefield tensor shape} & ---        & $\mathbf{(358,\, 619,\, 1000)}$      \\
        \textbf{Shot-gather cube shape} & ---        & $\mathbf{(64,\, 572,\, 1000)}$       \\
        \bottomrule
    \end{tabular}
\end{table}

Here $M$ is the spatial half-stencil width: the number of grid points on each side of the central node used to approximate the second spatial derivative. An 8th-order scheme ($M = 4$) reaches four neighbors in each direction, a common choice that balances accuracy against computational cost.

\paragraph{Absorbing boundary and slice margin.}
Each simulation grid is padded on its left, right, and bottom with a
60-cell sponge damping zone (\eqref{eq:cerjan_pde}, \eqref{eq:cerjan_damp}); the
top edge is a free surface \eqref{eq:free_surface}. The slice extraction step
(\cref{app:seismic_forward:slices}) reserves a 10\,\% lateral margin
from each cube edge, and source positions reserve an additional 0.5\,km
margin inside the absorbing-boundary region during forward modeling, so
sources radiate into the physical interior rather than into the damping
layer.

We generated the wavefields and shot-gather cubes for each of the five
frequency bands. The bands differ only in the source wavelet, so the
grid, time step, and saved tensor shape are identical across all bands.
The Nyquist margin in \cref{tab:bands} is the ratio of the highest
frequency the saved sampling can faithfully represent
($1/(2\dtsaved) \approx 35.7$\,Hz at $\dtsaved = 14$\,ms) to the band's
upper cutoff $\fmax$. A margin of $1\times$ would put $\fmax$ exactly at
the Nyquist limit; values larger than $1\times$ mean the band is
oversampled and aliasing-free at the saved rate. We chose $\dtsaved =
    14$\,ms so that all five bands clear Nyquist with a margin of at least
$\sim 1.4\times$.

\begin{table}[ht]
    \centering
    \caption{Frequency bands for wavefield and shot-gather generation.
        Nyquist margin = $1 / (2 \fmax \dtsaved)$ at $\dtsaved = 14$\,ms,
        i.e., the factor by which each band's upper cutoff falls below
        the saved-sampling Nyquist limit of $35.7$\,Hz. We use only the
        3--6\,Hz band for the wavefield prediction experiments in
        \cref{sec:wavefield_prediction_experiments}; all five bands are
        released on the \huggingface dataset.}
    \label{tab:bands}
    \smallskip
    \begin{tabular}{@{}cccc@{}}
        \toprule
        Band & Frequency range & $\fmax$ (Hz) & Nyquist margin \\
        \midrule
        1    & 3--6\,Hz        & 6            & 5.95$\times$   \\
        2    & 3--8.5\,Hz      & 8.5          & 4.20$\times$   \\
        3    & 3--12\,Hz       & 12           & 2.98$\times$   \\
        4    & 3--17.5\,Hz     & 17.5         & 2.04$\times$   \\
        5    & 3--25\,Hz       & 25           & 1.43$\times$   \\
        \bottomrule
    \end{tabular}
\end{table}

\subsection{Source Wavelet}
\label{sec:wavelet}

The source time function is a Ricker pulse with peak frequency $f_0$, then
bandpass-filtered with a 4th-order zero-phase Butterworth (forward-backward
filtering via \texttt{scipy.signal.sosfiltfilt}) at the band corners listed
in \cref{tab:bands}. The peak frequencies are
$f_0 \in \{4.50, 5.75, 7.50, 10.25, 14.00\}$\,Hz for the five bands
respectively, and each wavelet is centered at $t_0 = 1/f_0$.

After filtering, the amplitude of the wavelet depends sensitively on the
passband, so we apply a two-step normalization for numerical stability and
physical scaling. We first divide by the $\ell_2$ norm of the filtered
signal, which puts every band's wavelet on the same numerical footing. We
then multiply by $\sqrt{B / 24\,\mathrm{Hz}}$, where $B$ is the band's
bandwidth and $24$\,Hz is a fixed reference bandwidth chosen to match a
typical broadband seismic source (e.g.\ a 2--26\,Hz band). This
$\sqrt{B}$ scaling restores the Parseval relationship between bandwidth
and energy: a band-limited signal's radiated energy is proportional to
its passband width, so wider-band sources radiate more energy than
narrower-band sources, matching what a real source array does. Without
this rescaling, unit-energy normalization alone would make all five
bands radiate the same total energy, which is not physical.

\subsection{Wavefield and shot-gather acquisition geometry}
\label{sec:acquisition}

All simulations are run in Devito \citep{louboutin2019devito}, a
SymPy-based domain-specific language that JIT-compiles vectorized
OpenMP-parallel C from finite-difference stencils. Two acquisition
configurations are used, sharing the same grid, time step, and absorbing
boundary but differing in source count, receiver geometry, and recording
time.

\paragraph{Wavefield generation (5\,s, single source, no receivers).}
For each (slice, band) pair we run one 5\,s acoustic simulation with a
single source placed at depth 10\,m and a uniformly random horizontal
position drawn from the lateral interval $[1.1\,\mathrm{km},\,
            x_\mathrm{max} - 1.1\,\mathrm{km}]$. The 1.1\,km margin excludes the
60-cell sponge (0.6\,km) plus a 0.5\,km surveying buffer on each side, so
the source pulse propagates entirely inside the physical domain rather
than into the damping zone. Per-slice randomization is reproducible:
the seed is $42 \oplus \mathrm{CRC32}$(slice filename). We save the
full wavefield $p(\mathbf{x}, t)$ subsampled in time by $k = 14$,
producing one tensor per (slice, band) of shape $(358, n_x, 619)$.
With 4{,}276 slices and 5 bands, this gives 21{,}380 wavefield tensors
in total.

\begin{figure}[h]
    \centering
    \includegraphics[width=\linewidth]{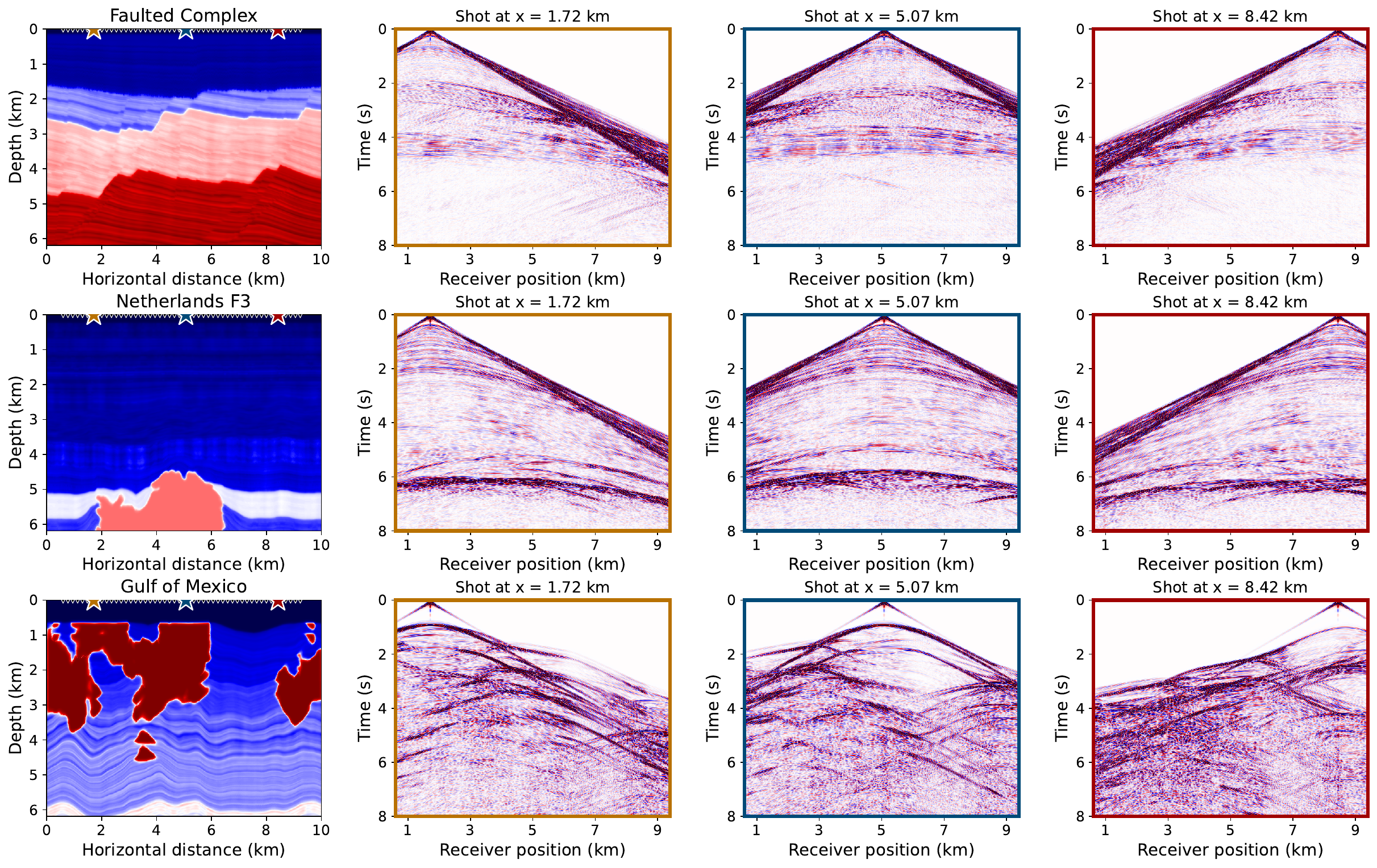}
    \caption{Representative 8\,s shot gathers from three training settings: Fault (top row), F3 (middle row), and GoM (bottom row).
        For each setting, the gathers correspond to three near-surface source positions on the left, center, and right of the lateral extent, color-coded to indicate source location.}
    \label{fig:shotgather_main}
\end{figure}

\paragraph{Shot-gather generation (8\,s, 64 sources, 1000 receivers).}
For each (slice, band) pair we run one 8\,s simulation that emulates a
marine streamer survey. 64 source positions are uniformly spaced across
the same valid lateral interval at depth 10\,m, and 1000 receivers at
depth 10\,m are uniformly spaced across the full lateral extent. The
receiver array is fixed in place (common-receiver streamer geometry):
the same 1000 receivers record all 64 source firings, rather than a
streamer that translates with the source. The output cube has shape
$(64, 572, 1000) = (\mathrm{sources}, \mathrm{time}, \mathrm{receivers})$,
where 572 is the number of saved time samples after $k = 14$
subsampling of the 8001 computational steps. We use 8\,s rather than
5\,s here so the deepest reflections from the 6.19\,km depth column
have time to return to the surface
($2 \cdot 6.19\,\mathrm{km} / v_\mathrm{shallow} \approx 8$\,s with
$v_\mathrm{shallow} \approx 1500$\,m/s in the water layer).
\cref{fig:shotgather_main} shows representative cubes from three
training settings.

\section{Wavefield Prediction}
\label{app:wavefield_prediction}

This appendix gives architecture, rollout, loss, training, and evaluation details for the wavefield-prediction experiments of \cref{sec:wavefield_prediction_experiments}.
The task is to predict the wavefield (5\,s of acoustic propagation at 3--6\,Hz) given the velocity slice and a binary source mask; we evaluate three operators: TFNO (chunked autoregressive forward prediction), TFNO-interp (anchor-bounded interior prediction), and DPOT (transformer with adaptive Fourier mixing).

\subsection{Task setup and preprocessing}
\label{app:wavefield_prediction:task}

\paragraph{Tensor shapes.}
Each chunked-rollout example has input shape $(C_\text{in}, H, W) = (12, 619, 1000)$ where
$C_\text{in} = T_\text{in} + 2$ stacks the 10 wavefield seed frames, the velocity
slice, and the binary source mask. The per-chunk target has shape
$(T_\text{out}, H, W) = (50, 619, 1000)$. The full forward-modeled trajectory
has 358 saved snapshots; the first 10 form the seed window, leaving 348 frames
to predict, so the inference-time rollout chains $\lceil 348/50 \rceil = 7$
chunks. The seventh chunk's leading 48 frames are kept and its trailing 2 are
discarded.

\paragraph{Normalization.}
Wavefield amplitudes vary by orders of magnitude across samples (different source
positions, velocity contrasts, recording windows), so each sample is normalized
by its own peak input amplitude:
\[
    s = \max\nolimits_{t, y, x} |x_\text{wf}(t, y, x)|, \quad
    \tilde{x}_\text{wf} = x_\text{wf} / s, \quad
    \tilde{y} = y / s.
\]
The same scale $s$ is applied to the target so the model learns a unit-amplitude
mapping. At inference time the prediction is rescaled by $s$ to recover absolute
amplitudes. Velocities use a global min--max to $[0, 1]$ over $[v_\text{min},
            v_\text{max}] = [1500, 5000]$~m/s; the source mask is already binary and is
passed through unchanged.

\subsection{Architectures}
\label{app:wavefield_prediction:architectures}

The three operators we compare are TFNO, TFNO-interp, and DPOT. TFNO and
TFNO-interp share the same channel-projection layout following
\citet{zhang2023elasticfno}: a $1\times 1$ convolution \texttt{Dp1} compresses
the $C_\text{in}$ input channels to a latent width $d_w$, the backbone
operates on $(d_w, H, W)$, and a $1\times 1$ convolution \texttt{Dp2}
expands the latent representation to the per-chunk output channel count.
DPOT does not use external \texttt{Dp1}/\texttt{Dp2} layers; its patch
embedding compresses the $C_\text{in}$ input channels directly into the
embedding space, and an unpatchify head produces the output frames.

\paragraph{TFNO.}
The Tucker-factorized Fourier Neural Operator
\citep{li2021fno, kossaifi2024tfno} replaces the dense complex spectral weight
tensor with a Tucker decomposition at rank fraction $0.5$, reducing the spectral
parameter count by ${\sim}4\times$ at the same width. The backbone configuration:
4 layers, $\texttt{n\_modes} = (32, 64)$, $\texttt{hidden\_channels} = 48$,
$d_w = 40$, GELU activations, linear FNO skips, soft-gating channel-MLP skip
connections, grid positional embedding, and 5\,\% per-side zero-padding on the
spatial domain to break the FFT's implicit periodic boundary (without padding
the absorbing-boundary energy at the bottom of the volume wraps to the free
surface). The $T_\text{out} = 50$ output channels of \texttt{Dp2} produce one
chunk per forward pass.

\paragraph{TFNO-interp.}
The interp variant uses the same TFNO2d backbone hyperparameters as TFNO but
wraps it in a \texttt{BoundaryAnchoredInterpPredictor}: instead of forward
prediction past a single seed window, it fills a $\tau = 20$ frame interior gap
between two stored 20-frame anchor windows ($T_L = T_R = 20$). The wrapper
predicts a $\tau$-frame residual via a sine-windowed head:
\[
    \hat u(t) = (1-\alpha_t)\, u_\text{anchor end}
    + \alpha_t\, u_\text{anchor start}
    + \sin(\pi \alpha_t)\, r_\theta(t),
    \qquad \alpha_t = \tfrac{t+1}{\tau+1},
\]
where the linear-in-$\alpha_t$ part interpolates between the anchor frames,
$r_\theta(t)$ is the network output, and the sine envelope $\sin(\pi \alpha_t)$
forces the prediction to match the anchor frames exactly at the gap endpoints
($\alpha_t=0,\,1$) while peaking in the middle of the gap.

\paragraph{DPOT.}
DPOT \citep{hao2024dpot} is a ViT-style operator with patch embedding, AFNO
(Adaptive Fourier) mixing, and learned positional information. We use the
\texttt{DPOTAdapterSimple} configuration with embedding dimension 256, 4 transformer
blocks, $\texttt{num\_blocks} = 4$ (block-diagonal AFNO structure),
$\texttt{modes} = 16$ (AFNO frequency truncation), MLP ratio 1.0, patch size
8, and GELU activations. As noted above, DPOT projects the $C_\text{in}$
input channels into the embedding space through its patch embedding rather
than through external $1\times 1$ convolutions, and an unpatchify head
produces the $T_\text{out}$ output frames.

\subsection{Hybrid (chunked autoregressive) rollout}
\label{app:wavefield_prediction:rollout}

The full forward-modeled trajectory has 358 saved snapshots
(\cref{app:seismic_forward}); the model only predicts $T_\text{out} = 50$ at
once. Autoregressive chaining is the operator-learning analogue of how a
standard finite-difference solver advances from the $t = 0$ initial
condition to all subsequent times: each chunk uses the model's previous
output as its next initial condition, so the same single trained operator
generates the full trajectory rather than a separate model per horizon.
Two extreme rollout strategies are unattractive: single-step
($T_\text{out} = 1$) autoregression accumulates error over 348 steps and is
known to compound to large drift in long horizons; full-trajectory
($T_\text{out} = 348$) prediction in one forward pass would inflate the
output channel count by $\sim$7$\times$ relative to our chunk
size, exceeding the activation budget at the field-scale grid of
$(619, 1000)$ on a single A100. We instead use a chunked autoregressive
rollout (\cref{alg:rollout}): the model produces $T_\text{out}$ frames per
pass, the last $T_\text{in}$ predicted frames seed the next pass, and
per-chunk amplitude renormalization mirrors the per-sample normalization
used during training. With $T_\text{out} = 50$ and 348 frames to predict,
$\lceil 348/50 \rceil = 7$ chunks suffice; only the leading 48 frames of
the seventh chunk are retained (the trailing 2 fall past the end of the
saved trajectory and are discarded).

\begin{algorithm}[h]
    \caption{\texttt{autoregressive\_rollout} --- chunked wavefield prediction.}
    \label{alg:rollout}
    \begin{algorithmic}[1]
        \Require Trained model $M$; seed $u_{0:T_\text{in}}$; velocity $v$; source mask $m$; total length $T$; chunk size $T_\text{out}$; bounds $v_\text{min}, v_\text{max}$.
        \Ensure Predicted wavefield $\hat u_{T_\text{in}:T}$.
        \State Normalize velocity once: $\tilde v \gets (v - v_\text{min}) / (v_\text{max} - v_\text{min})$.
        \State Buffer $b \gets u_{0:T_\text{in}}$ (sliding window of last $T_\text{in}$ frames).
        \For{$k = 0, 1, \ldots, \lceil (T - T_\text{in}) / T_\text{out} \rceil - 1$}
        \State $s \gets \max |b|$; $\tilde b \gets b / s$. \Comment{per-chunk amplitude scale}
        \State $x \gets [\tilde b,\ \tilde v,\ m]$ stacked on the channel axis.
        \State $\hat u_\text{chunk} \gets s \cdot M(x)$. \Comment{denormalize the prediction}
        \State Append the needed prefix of $\hat u_\text{chunk}$ to the output.
        \State Update $b$: take the last $T_\text{in}$ frames of $\hat u_\text{chunk}$ (or slide if $T_\text{out} < T_\text{in}$).
        \EndFor
    \end{algorithmic}
\end{algorithm}

\subsection{Loss}
\label{app:wavefield_prediction:loss}

Training minimizes a temporally-weighted relative $L^2$ loss:
\begin{equation*}
    \mathcal{L} = \mathbb{E}_\text{batch}\left[
        \sqrt{\tfrac{\sum_t w(t)\,\|\hat u_t - u_t\|_2^2}{\sum_t w(t)\,\|u_t\|_2^2 + \varepsilon}}
        \right].
\end{equation*}

The temporal weight schedule $w(t)$ is motivated by error compounding in
autoregressive rollouts: errors in the first few predicted frames seed the
input buffer for the next chunk and propagate forward, so accuracy at
small $t$ has higher leverage on long-horizon stability than accuracy at
large $t$. \citet{coker2025stlw} formulate the same hypothesis for
autoregressive neural operators on 1D PDEs (``giving higher priority to
earlier timesteps will reduce the overall propagation of errors and enhance
training stability''). Our schedule applies the practice statically:
$w(t)$ decays linearly from $w_\text{start}=1.0$ at $t=0$ to
$w_\text{end}=0.5$ at $t=T_\text{out}-1$, scaling the gradient signal in
proportion to that leverage.

TFNO and DPOT use the linear temporal weighting just described. TFNO-interp
uses uniform temporal weighting ($w_\text{start} = w_\text{end} = 1.0$)
because its prediction is bounded symmetrically by anchor frames at both
endpoints
--- there is no early-vs-late asymmetry to exploit --- plus an additional
first-difference temporal-derivative term, weight $\alpha_\text{dt}=0.1$,
that penalizes $\|\partial_t (\hat u - u)\|_2^2$ across the gap.

\subsection{Training configuration}
\label{app:wavefield_prediction:training}

\begin{table}[h]
    \centering
    \small
    \caption{Wavefield prediction training hyperparameters per architecture.
        The three runs share the
        same optimizer, schedule shape, weight decay, gradient clipping,
        mixed-precision policy, batch size, and epoch count, so the
        comparison is on architecture rather than training hyperparameters.
        The one deviation is the base learning rate (see text).}
    \label{tab:wavefield_training}
    \begin{tabular}{lccc}
        \toprule
                                       & TFNO     & TFNO-interp & DPOT     \\
        \midrule
        Optimizer                      & AdamW    & AdamW       & AdamW    \\
        Learning rate                  & 2e-3     & 2e-3        & 5e-4     \\
        Weight decay                   & 1e-4     & 1e-4        & 1e-4     \\
        Schedule                       & cosine   & cosine      & cosine   \\
        Min LR (cosine floor)          & 1e-5     & 1e-5        & 1e-5     \\
        Gradient clip ($\ell_2$)       & 1.0      & 1.0         & 1.0      \\
        Mixed precision                & yes      & yes         & yes      \\
        Per-GPU batch                  & 8        & 8           & 8        \\
        Grad-accumulation steps        & 4        & 4           & 4        \\
        Effective batch                & 64       & 64          & 64       \\
        Epochs                         & 10       & 10          & 10       \\
        $w_\text{start}, w_\text{end}$ & 1.0, 0.5 & 1.0, 1.0    & 1.0, 0.5 \\
        $\alpha_\text{dt}$             & 0.0      & 0.1         & 0.0      \\
        \bottomrule
    \end{tabular}
\end{table}

As shown in \cref{tab:wavefield_training}, the one training-hyperparameter deviation across architectures is the base
learning rate: in preliminary short runs DPOT exhibited substantially larger
batch-level oscillations in training loss at $2 \times 10^{-3}$, so we
lowered DPOT's base LR to $5 \times 10^{-4}$ with the same cosine decay.
The adjustment is consistent with the original DPOT recipe
\citep{hao2024dpot}, which uses AdamW with a one-cycle schedule and a long
warmup phase, implying a much smaller effective learning rate at the onset
of training; using a smaller cosine peak LR for DPOT mirrors that effect
within our cosine-only schedule.

All three runs are trained on $2 \times$ NVIDIA A100 80\,GB via DDP, with effective batch 64 ($8 \times 4$ grad-accum steps
$\times 2$ GPUs). We decompress the source HDF5 files to NPY files
during preprocessing and use the NPY files for training, which allows us to
avoid the decompression overhead of HDF5. All reported
runs use a single seed.
Each run takes about 24 hours.

\subsection{Evaluation pipeline and reported metrics}
\label{app:wavefield_prediction:eval}

We evaluate trained operators on the in-distribution test set (100 slices,
five training settings with non-overlapping indices) and the
out-of-distribution test set (80 Penobscot slices). For each (operator,
slice) pair we run a forward rollout from the seed window, compute
per-sample per-frame and per-wavenumber metrics, write the per-sample
numbers into a parquet manifest, and aggregate across the test set to
produce the figures reported in
\cref{sec:wavefield_prediction_experiments}.

The metrics computed per sample are: relative $L^2$ error per frame
$L^2_\text{RE}(t) = \|\hat u_t - u_t\|_2 / (\|u_t\|_2 + \varepsilon)$
(plotted on the log-y ``L2RE'' axis of \cref{fig:l2_vs_time} as a per-frame
mean across the test set, on a physical-time axis, with in-distribution and
out-of-distribution test splits side-by-side); a per-wavenumber relative
power-spectrum error
$|P_\text{pred}(k) - P_\text{gt}(k)| / (P_\text{gt}(k) + \varepsilon)$,
where $P(k)$ is the azimuthally-averaged radial 2D power spectrum first
averaged over the trajectory's frames (this is \emph{not} a vector $L^2$
norm; it is plotted as ``Relative spectral error'' in \cref{fig:spectral_l2});
trajectory-wide rel-$L^2$; max-absolute-error; energy ratio
$E[-1]/E[0]$ and per-frame energy std; divergence time (first frame where
$\|\hat u\|_\infty > 10\,\|u\|_\infty$ or NaN/Inf); per-frame SSIM; and a
physics-residual ratio computed from the discretized acoustic wave operator.

The signed-error panel grids show $\hat u - u$ at four selected frames on
a shared symmetric color scale, with a velocity-grayscale backdrop
($\alpha = 0.3$) and a star marking the source position. The
chunked-rollout panel (\cref{fig:chunk_grid}) is a one-shot teacher-forced
chunk seeded from GT frames $[240, 250)$ and predicting frames $[250, 300)$
at sampled positions 256, 270, 285, 299 (no autoregressive chaining for
this figure). The TFNO-interp panel (\cref{fig:gt_vs_interp}) shows signed
error at gap centers (frames 30, 150, 230, 270) of the stitched rollout;
its colorbar reuses \cref{fig:chunk_grid}'s clim so error magnitudes are
directly comparable across the two protocols. \cref{fig:l2_vs_time}
additionally shades TFNO-interp's gap-prediction windows in light gray and
diagonally hatches the trailing window past the last anchor (frames
340--358, equivalently 4.76--5.00\,s, which the model does not predict).

Each metric tells us a different thing: the per-frame rel-$L^2$ curve
shows how prediction error compounds (or stays bounded) as the rollout
extends; the spectral curve isolates the failure mode of neural operators
at high wavenumbers, the spectral-bias regime; the energy ratio diagnoses
whether the rollout is dissipating or blowing up; divergence time picks
out catastrophic failures; and the physics-residual ratio measures how
well the prediction satisfies the governing PDE without referring to the
ground truth.

\subsection{Per-wavenumber and per-geology breakdowns}
\label{app:wavefield_prediction:breakdowns}

\begin{figure}[t]
    \centering
    \includegraphics[width=\linewidth]{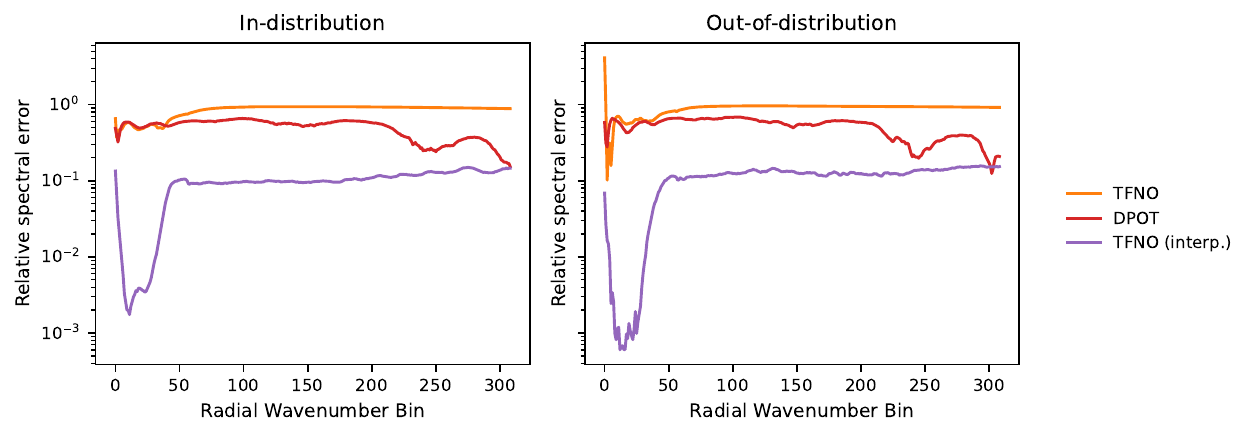}
    \caption{Per-wavenumber relative spectral error on the in-distribution (left) and out-of-distribution (right) test sets.
        TFNO-interp dominates across the spectrum; DPOT outperforms TFNO at high wavenumbers.}
    \label{fig:spectral_l2}
\end{figure}

\paragraph{Per-wavenumber error.}
\cref{fig:spectral_l2} plots the per-wavenumber relative power-spectrum error.
TFNO-interp dominates across the spectrum; DPOT outperforms TFNO at high wavenumbers.

\begin{figure}[t]
    \centering
    \includegraphics[width=\linewidth]{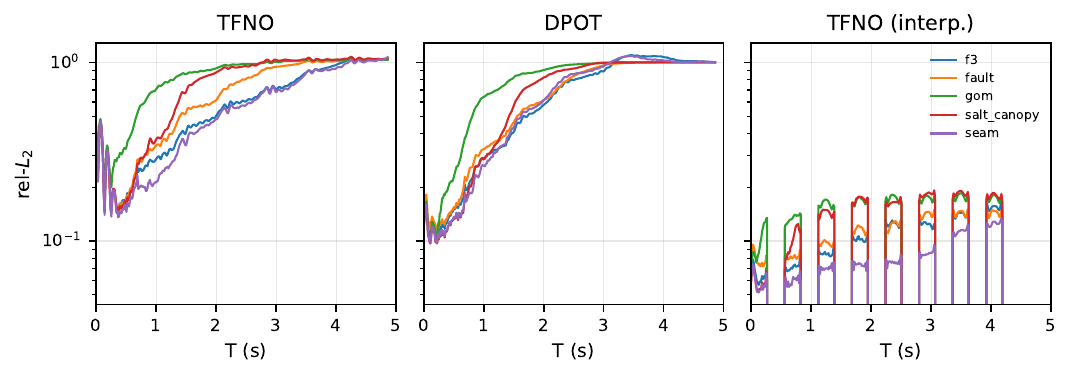}
    \caption{L2RE versus time for each geological setting in the in-distribution test set.
        Gulf of Mexico and Salt Canopy are the most challenging settings, while SEAM and F3 are the easiest.}
    \label{fig:per_geology}
\end{figure}

\paragraph{Per-geology error.}
\cref{fig:per_geology} displays the L2RE per geological setting in the in-distribution test set.
Across all architectures, the Gulf of Mexico and Salt Canopy settings are the most challenging, while SEAM and F3 are the easiest.
We suspect this is due to Gulf of Mexico and Salt Canopy being the most salt-rich settings in the dataset.

\section{End-to-End Seismic Inversion}
\label{app:inversion}

This appendix gives architecture, loss, and training details for the end-to-end
encoder--decoder inversion experiments of \cref{sec:inversion_experiments}. The
task is to map a 3D shot-gather cube (produced by 8 s of acoustic propagation at
3--25 Hz) to a 2D $v_p(\mathbf{x})$ velocity model.

\subsection{Task setup and preprocessing}
\label{app:inversion:task}

\paragraph{Tensor shapes.}
The input is a 3D shot-gather cube of shape $(n_\text{shots}, n_\text{time},
    n_\text{receivers}) = (64, 572, 1000)$, presented to the network as
$(B, 1, 64, 572, 1000)$, where $B$ is the batchsize.
The output is a 2D velocity slice of shape $(B, 1, 619, 1000)$ in the normalized velocity range $[0, 1]$.

\paragraph{Normalization.}
Shot gathers are normalized by sign-preserving clipping.
We compute a single global clipping threshold once during preprocessing.
We randomly sample 200 training shot-gather cubes, take the 98th percentile of the absolute
amplitudes within each cube, then take the 98th percentile of those per-cube
values to obtain the global threshold.
At training time, every shot-gather cube has its amplitudes clipped at plus or minus this threshold and divided by the threshold, mapping values to $[-1, 1]$.
Velocities are mapped to $[0, 1]$ by an affine transformation that sends
1500~m/s to $0$ and 5000~m/s to $1$.

\paragraph{Augmentation.}
During training, the (shot-gather cube, velocity slice) pairs are augmented
with a horizontal flip (probability $0.5$, applied jointly to the receiver
axis of the cube and the horizontal axis of the velocity slice so their relationship through the acoustic wave equation \eqref{eq:acoustic} is preserved) and additive Gaussian noise with signal-to-noise ratio drawn uniformly between $[10, 30]$ dB.

\subsection{Architectures}
\label{app:inversion:architectures}

We compare three encoder--decoder architectures.
The Transformer and CNN encoders share an identical architecture for the decoder.
InversionNet is used because it has been employed in OpenFWI \citep{deng2022openfwi}.

\paragraph{InversionNet.}
A direct port of the 2D encoder--decoder of \citet{wu2020inversionnet}. The
shot-gather cube is treated as a 2D image with 64 channels (one per shot)
and processed by a convolutional encoder followed by a deconvolution
(a.k.a. transposed convolution) decoder, with a final resampling to the
$(619, 1000)$ output grid.

\paragraph{3D residual CNN encoder + 2D decoder.}
A 3D residual encoder following nnU-Net ResEnc principles
\citep{isensee2024nnunetrevisited}. The first encoder stage uses an
asymmetric stride that aggressively downsamples the receiver axis while
preserving the shot axis; remaining stages use symmetric strides.
A 3D-to-2D transition then collapses the shot dimension to produce a 2D
feature map, which is fed to a 2D residual decoder that lifts the
feature map back to the $(619, 1000)$ output grid and applies a sigmoid to
yield a single normalized velocity channel.

\paragraph{3D Swin encoder + 2D decoder.}
A small CNN stem performs spatial downsampling of the cube.
A 3D patch embedding then maps the result to a token grid that is processed by a
3D Swin encoder \citep{liu2021swin, hatamizadeh2022swinunetr}
via windowed self-attention with shifted-window blocks and patch merging
between stages.
The output of the encoder is collapsed to a 2D feature map and
fed to the same decoder architecture used by the CNN architecture.

\subsection{Loss}
\label{app:inversion:loss}

Training minimizes the mean squared error between the predicted and
ground-truth normalized velocity slices, averaged over the training set and
the height and width of each slice:
\[
    \mathcal{L} = \frac{1}{N\,H\,W} \sum_{n=1}^{N} \sum_{i=1}^{H} \sum_{j=1}^{W}
    \left( \hat v_{n, i, j} - v_{n, i, j} \right)^2,
\]
where $N$ is the number of training pairs, $H = 619$ and $W = 1000$ are the
height and width of the velocity slice, $v_{n, i, j} \in [0, 1]$ is the
normalized ground-truth velocity at pixel $(i, j)$ of the $n$th training
sample, and $\hat v_{n, i, j} \in [0, 1]$ is the corresponding model
prediction.

\subsection{Training configuration}
\label{app:inversion:training}

\begin{table}[h]
    \centering
    \small
    \caption{Inversion training hyperparameters per architecture
        (3--25 Hz frequency band, 8 s propagation). All three architectures
        share the same optimizer, schedule, and augmentation.}
    \label{tab:inversion_training}
    \begin{tabular}{lccc}
        \toprule
                                   & InversionNet & CNN        & Swin       \\
        \midrule
        Optimizer                  & AdamW        & AdamW      & AdamW      \\
        Learning rate              & 1e-4         & 1e-4       & 1e-4       \\
        Weight decay               & 1e-4         & 1e-4       & 1e-4       \\
        Schedule                   & cosine       & cosine     & cosine     \\
        Min LR (cosine floor)      & 1e-6         & 1e-6       & 1e-6       \\
        Gradient clip ($\ell_2$)   & 1.0          & 1.0        & 1.0        \\
        Mixed precision            & yes          & yes        & yes        \\
        Per-GPU batchsize          & 8            & 8          & 8          \\
        Grad-accumulation steps    & 1            & 1          & 1          \\
        Effective batchsize        & 16           & 16         & 16         \\
        Epochs                     & 100          & 100        & 100        \\
        Horizontal flip ($p$)      & 0.5          & 0.5        & 0.5        \\
        Noise injection range (dB) & $[10, 30]$   & $[10, 30]$ & $[10, 30]$ \\
        \bottomrule
    \end{tabular}
\end{table}

We list training hyperparameters per architecture in \cref{tab:inversion_training}.
All three architectures train on $2 \times$ NVIDIA A100 80 GB via PyTorch
DDP, with effective batchsize 16 ($8 \times 1$ grad-accum steps $\times 2$
GPUs). We train three random seeds per architecture and report mean RMSE
and SSIM in the main text. We decompress the source HDF5 files to NPY files
during preprocessing and use the NPY files for training, which allows us to
avoid the decompression overhead of HDF5.
Each training run takes about 20 hours.



\end{document}